\documentclass[pdflatex,sn-mathphys-num]{sn-jnl}

\usepackage{graphicx}%
\usepackage{multirow}%
\usepackage{amsmath,amssymb,amsfonts}%
\usepackage{amsthm}%
\usepackage{mathrsfs}%
\usepackage[title]{appendix}%
\usepackage{xcolor}%
\usepackage{textcomp}%
\usepackage{manyfoot}%
\usepackage{booktabs}%
\usepackage{algorithm}%
\usepackage{algorithmicx}%
\usepackage{algpseudocode}%
\usepackage{listings}%
\usepackage[most]{tcolorbox}
\usepackage{listings} 
\usepackage{amsmath}
\usepackage{fvextra}
\usepackage[normalem]{ulem}
\useunder{\uline}{\ul}{}
\usepackage{multirow}
\DefineVerbatimEnvironment{verbatim}{Verbatim}{breaklines}
\usepackage[normalem]{ulem}
\usepackage{hyperref}

\usepackage{booktabs}
\usepackage{subcaption} 
\usepackage{array} 
\usepackage{amsmath}

\usepackage{etoolbox}

\theoremstyle{thmstyleone}%
\theoremstyle{thmstyletwo}%

\theoremstyle{thmstylethree}%

\begin{document}

\newtcolorbox{promptbox}[1][]{
  breakable,
  title=#1,
  colback=gray!5,
  colframe=black,
  colbacktitle=gray!15,
  coltitle=black,
  fonttitle=\bfseries,
  bottomrule=1.5pt,
  toprule=1.5pt,
  leftrule=1pt,
  rightrule=1pt,
  arc=0pt,
  outer arc=0pt,
  enhanced,
  before upper={\parindent=1.5em} 
}

\newlength{\panelwidth}
\newlength{\gapwidth}
\newlength{\singlepanelwidth}

\title{Large Language Models are Near-Optimal Decision-Makers with a Non-Human Learning Behavior}


\makeatletter
\patchcmd{\@maketitle}{Contributing authors:\ }{}{}{}
\makeatother

\author[a, †]{Hao Li}
\author[b, †]{Gengrui Zhang}
\author[c,d]{Petter Holme}
\author[e, ‡]{Shuyue Hu}
\author[a, ‡]{Zhen Wang}
\affil[a]{School of Cybersecurity, Northwestern Polytechnical University, China}
\affil[b]{Department of Psychology, University of Southern California, United States}
\affil[c]{Department of Computer Science, Aalto University, Finland}
\affil[d]{Center for Computational Social Science, Kobe University, Japan}
\affil[e]{Shanghai Artificial Intelligence Laboratory, China}
\email{\centering
li.hao@mail.nwpu.edu.cn; gengruiz@usc.edu; petter.holme@aalto.fi\\ 
hushuyue@pjlab.org.cn; w-zhen@nwpu.edu.cn}

\let\thefootnoteorig\thefootnote
\def\thefootnote{\arabic{footnote}} 
\def\thefootnote{†}\footnotetext{These authors contributed equally; this work was partly done during their internship at Shanghai Artificial Intelligence Laboratory.}
\def\thefootnote{‡}\footnotetext{Corresponding authors.}
\let\thefootnote\thefootnoteorig


\abstract{Human decision-making belongs to the foundation of our society and civilization, but we are on the verge of a future where much of it will be delegated to artificial intelligence. The arrival of Large Language Models (LLMs) has transformed the nature and scope of AI-supported decision-making; however, the process by which they learn to make decisions, compared to humans, remains poorly understood. In this study, we examined the decision-making behavior of five leading LLMs across three core dimensions of real-world decision-making: uncertainty, risk, and set-shifting. Using three well-established experimental psychology tasks designed to probe these dimensions, we benchmarked LLMs against 360 newly recruited human participants. Across all tasks, LLMs often outperformed humans, approaching near-optimal performance. Moreover, the processes underlying their decisions diverged fundamentally from those of humans. On the one hand, our finding demonstrates the ability of LLMs to manage uncertainty, calibrate risk, and adapt to changes. On the other hand, this disparity highlights the risks of relying on them as substitutes for human judgment, calling for further inquiry.}


%
%
%




\maketitle

Decision-making is a unifying theme of almost all the social and behavioral sciences. Moreover, it has been closely tied to artificial intelligence (AI), ever since the inception of the latter. It is no coincidence that one of AI's founders, Herbert Simon, began his career studying organizational decision-making~\cite{simon}. Simon once pointed out that ``the capacity of the human mind for formulating and solving complex problems is very small compared with the size of the problems whose solution is required for objectively rational behavior in the real world''~\cite{simon}. And so the first generation of AI built decision systems~\cite{turban1986integrating} by implementing logic and circumstantial knowledge. In contrast, the last decade of AI development has had a very different driving force---to imitate human intellectual output in broad generality. By training self-supervised artificial neural networks on immense textual corpora---so-called Large Language Models (LLMs)---it is commonly accepted that AI has now passed the Turing test~\cite{turing_test,mei2024turing}. LLMs do make decisions if we prompt them to do so, but are those decisions made with a capacity as ``very small'' as the humans they try to emulate? That is the central question we pursue in this paper.

LLMs are remarkably versatile, capable of generating meaningful output across a wide range of tasks, and are, unsurprisingly, already being deployed in real-world settings to make or influence decisions~\cite{bommasani2021opportunities,handler2024large,chensurvey}.
For a concrete example, in a legal case in Colombia concerning the medical expenses of an autistic boy, the judge not only queried ChatGPT, asking ``Is an autistic minor exonerated from paying fees for their therapies?'', but also cited both the prompt and the answer in the official ruling~\cite{Taylor2023ChatGPTRuling}.
Likewise, LinkedIn, a major professional networking platform, has deployed an LLM-based chatbot to assist recruiters in automating critical hiring decisions, such as the shortlisting of job candidates, thereby directly shaping individuals' career opportunities~\cite{AlbaYin2024GenAIHiring}.
Globally, surveys estimate that 71\% of major organizations have already regularly used generative AI in at least one business function~\cite{singla_state_2025}. 
For better or worse, generative AI is rapidly becoming integral to decision-making infrastructures, where each decision made, like those made by humans, has the power to shape individual lives and, potentially, ripple outward to influence entire societies~\cite{noy2023experimental,extance2023chatgpt,williams2024evaluating,goh2025gpt}.

However, are generative AI systems truly ready to assist or even replace humans in decision-making?  Recent studies have reported strong performance for the decision-making of these systems in domains such as healthcare~\cite{singhal2023large,hager2024evaluation},  finance~\cite{chen-etal-2021-finqa,xie2024finben}, and strategic games~\cite{wangvoyager,tan2024cradle,akata2025playing,wang2024large}. Yet, arguably more critical---but far less explored---is how they arrive at these decisions, particularly when operating outside the narrowly defined scenarios. Domain-specific tasks, though valuable, often introduce numerous confounds, such as background knowledge, ethical considerations, and cultural biases, which can obscure the working mechanisms that drive system behaviors. Moreover, although these tasks capture real-world complexity, they often conflate conceptually distinct dimensions of decision-making, such as risk and uncertainty. For instance, a medical diagnosis benchmark may fold epistemic uncertainty about a patient's true condition into the known risks of various treatments~\cite{hager2024evaluation}. Consequently, even strong performance on these tasks offers little diagnostic insight into whether the system effectively excels at managing uncertainty, calibrating risk, or both, let alone how these dimensions are reasoned about. In contrast, what fundamentally characterizes human decision-making capacity and their agency is not merely the correctness of a task-specific choice, but the ability to navigate different decision dimensions while simultaneously aligning their choices with goals~\cite{johnson2010decision,hastie2010rational,einhorn1981behavioral}.

In this study, we compare the decision-making of generative AI systems and humans, particularly under conditions of uncertainty, risk, or set-shifting. Here, decision-making refers to the process by which individuals assess multiple alternatives and choose actions that align with defined goals.
Uncertainty involves acting with incomplete information and ambiguous future outcomes, requiring a balance between short-term and long-term consequences~\cite{knight1921risk,camerer1992recent}. 
Risk entails evaluating potential gains and losses based on known probabilities, demanding careful judgment of outcome likelihoods~\cite{machina1987decision, kahneman2013prospect}. 
Set-shifting refers to dynamic environments where conditions evolve over time, requiring the ability to adapt strategies as new information emerges~\cite{brehmer1992dynamic,payne1993adaptive}.
Uncertainty, risk, and set-shifting are three key dimensions that dominate most real-world decisions~\cite{kochenderfer2018decision,ruggeri2020replicating,uddin2021cognitive}. They each represent a conceptually distinct construct, yield their own behavioral patterns, and engage their own cognitive mechanism~\cite{strategy1997deciding,sacre2019risk,konishi1998transient}.

Our methodology for investigating if and how generative AI systems navigate these decision dimensions aligns with recent calls for machine psychology~\cite{binz2023using,hagendorff2023machine,kosinski2024evaluating,hagendorff2024deception,chen2023emergence,lehr2025kernels}, which advocates the use of experimental psychology paradigms.  
Unlike domain-specific tasks, psychological tests, originally designed for humans, strip away contextual confounds and isolate distinct decision dimensions.  As a result, they can provide simplified yet tight experimental control, enable precise theory testing, and reveal general cognitive mechanisms that may extend across diverse domains.
We adapted three well-established psychological tests: the Iowa Gambling Task for uncertainty~\cite{bechara1994insensitivity}, the Cambridge Gambling Task for risk~\cite{rogers1999dissociable}, and the Wisconsin Card Sorting Task for set-shifting~\cite{berg1948simple} (see Methods for task details). 
We treated LLMs as subjects in these tests. 
To mitigate the potential memorization effects of these systems~\cite{hagendorff2023machine}, we reworded task descriptions and redesigned payoff structures while preserving the essence of the original tests.

We considered five leading LLMs: GPT (\textit{gpt-4o-2024-08-06}), \textcolor{black}{GPTo4m (\textit{o4-mini-2025-04-16})}, Claude (\textit{claude-3-5-sonnet-20240620}), Gemini (\textit{gemini-1.5-pro-002}), and \textcolor{black}{DeepSeek (\textit{DeepSeek-R1-2025-01-20})}. 
To provide a yardstick for their decision-making, we benchmarked them with 360 newly recruited human participants (120 per task), presenting both groups identical experimental instructions (see Methods, and \href{SI.com}{\textit{SI Appendix, Supplementary Note 1}} for details). Across all three tests, we consistently observed that LLMs were able to perform significantly better than human participants; however, the ways in which they arrived at their decisions were fundamentally different from those of humans. This key finding offers direct evidence of these systems’ general decision-making competence---particularly in managing uncertainty, calibrating risk, and adapting to change. On the other hand, it cautions against using these systems as stand-ins for human decision-making in real-world contexts or behavioral research~\cite{abdurahman2024perils,wang2025large,grossmann2023ai}. More broadly, it highlights the critical need for policymakers and system designers to carefully consider how much autonomy to delegate to such systems, ensure transparent communication with end-users about these systems' potential cognitive differences, and maintain meaningful human oversight, especially in domains where human-like reasoning and judgment are essential.


\section*{Results}
In each test, most evaluated LLMs outperformed human participants and approached near-optimal performance, yet relied on decision-making strategies that were notably different from those of humans. Computational models revealed that the posterior parameter estimates for LLMs diverged significantly from those of human participants, highlighting fundamental differences in underlying cognitive processes.

\begin{figure*} [!t]
       \centering
       \includegraphics[width=0.9\linewidth]{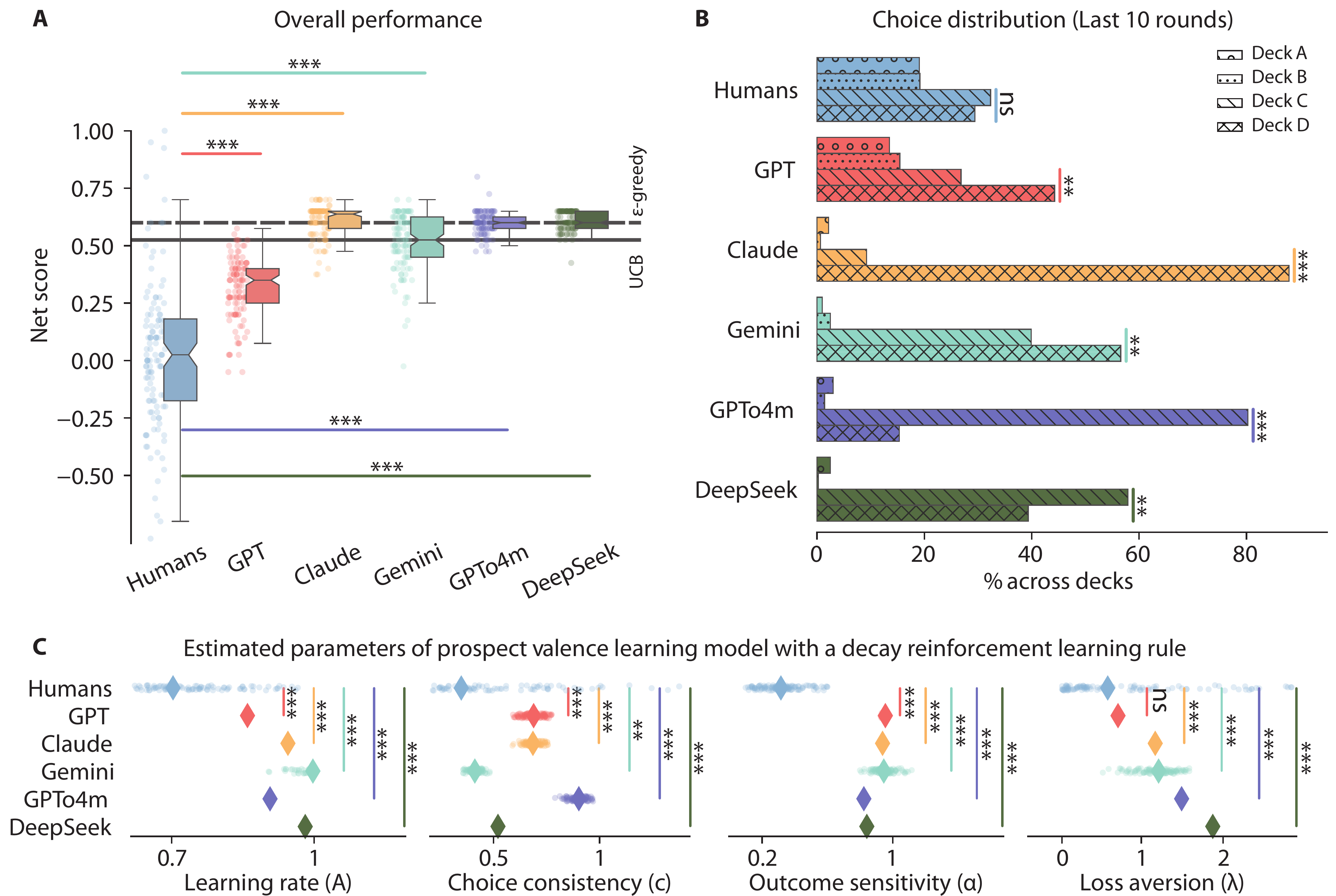}
       \vspace{1em}
       \caption{\textbf{All the LLMs significantly outperformed humans in the Iowa Gambling Task, yet differed in choice preferences and exhibited distinct parameter estimates in the prospect valence learning model compared to humans.} \textbf{Panel (A)} shows the net scores of LLMs, human participants, and two well-established strategies (Upper Confidence Bound and $\epsilon-$greedy).  A Mann-Whitney U test indicated that GPT-4o, GPTo4m, Claude, Gemini, and DeepSeek significantly outperformed human participants. Specifically, Claude achieved the highest median net score and GPTo4m showed the lowest variance.
       \textbf{Panel (B)} shows the distributions of deck selections over the last 10 rounds. Both human participants and LLMs predominantly favored the advantageous decks (C and D). 
       However, a two-proportion Z test revealed that GPT, Claude, and Gemini showed a stronger preference for deck D over deck C, whereas GPTo4m and DeepSeek showed the opposite preference.
       In contrast, human participants selected decks C and D in similar proportions. 
       \textbf{Panel (C)} presents the posterior estimates of the parameters in the prospect valence learning model. Mann-Whitney U tests indicated that compared to humans, all the LLMs demonstrated 
       higher learning rates ($A$), 
       greater sensitivity ($\alpha$) to outcomes, 
       and more tendency ($c$) to make deterministic decisions. 
       Moreover, except for GPT, they also showed stronger reactions ($\lambda$) to penalties than humans. GPT showed no significant difference from humans on the parameter of loss aversion. 
       All inferential (Mann–Whitney U tests) and descriptive statistics (means, medians, and standard deviations) for the Iowa Gambling Task are reported in \href{SI.com}{\textit{SI Appendix}, Table~\ref{tab:igt_results}}, including pairwise comparisons (panels a–c) and parameter estimates (panel d).
       All parameter estimates demonstrated satisfactory convergence with R-hat ($\hat{R}$) values below 1.01.}
       \label{fig:IGT_fig1}
\end{figure*}

\subsection*{Decision-making under uncertainty} 
The Iowa Gambling Task tested whether players can prioritize long-term benefits over short-term gains in an uncertain environment. During the task, participants repeatedly selected cards from four decks (labeled A, B, C, and D).  
Each choice offered an immediate reward, with the possibility of an occasional penalty. 
Unbeknownst to players, decks A and B were disadvantageous, offering higher immediate rewards but lower long-term expected payoffs, whereas decks C and D were advantageous, providing smaller immediate rewards but greater long-term gains.
The outcomes of individual card selections were unpredictable, requiring participants to infer the underlying payoff structure through repeated trials and feedback.

We measured the overall performance of human participants and LLMs using net scores~\cite{bull2015decision}, defined as the difference in the proportion of selections between the advantageous and disadvantageous decks.
Fig.~\ref{fig:IGT_fig1}A compares the net scores of human participants, two GPT models, Claude, Gemini, and DeepSeek. The five types of LLMs all significantly outperformed human participants, demonstrating higher task proficiency.
Among the models, Claude achieved the highest median net score, indicating better overall task performance, whereas GPTo4m showed the lowest variance, reflecting greater consistency across the task.
Additionally, when benchmarked against two well-established strategies for decision-making under uncertainty, namely, upper confidence bound~\cite{auer2002finite} and $\epsilon-$greedy~\cite{watkins1989learning}, Claude, Gemini, GPTo4m, and DeepSeek approached their near-optimal performance.
Linear regression models further revealed that, over time, these LLMs also learned faster than humans, demonstrating steeper slopes in their proportion of advantageous deck selections (\href{SI.com}{\textit{SI Appendix}, Fig.~\ref{fig:IGT_fig1_b}}). Thus, compared to humans, LLMs managed to identify advantageous decks earlier and adapted their choices more effectively to the task's reward-penalty structure.

Beyond performance, we analyzed the distribution of choices across decks to assess differences in decision strategies. As illustrated in Fig.~\ref{fig:IGT_fig1}B, during the final \textcolor{black}{ten} rounds, both human participants and LLMs predominantly selected the two advantageous decks (i.e., C and D), which have equivalent long-term expected returns. However, their choice distributions differed substantially. Human participants selected decks C and D in similar proportions, consistent with prior experimental findings in the literature~\cite{bechara1994insensitivity}. In contrast, LLMs exhibited systematic but heterogeneous preferences across the advantageous decks, with some models showing a stronger inclination toward deck D, which was associated with less frequent penalties, and others favoring deck C, which had more frequent but smaller penalties. This suggests that LLM behavior is more sensitive to penalty frequency, whereas human choices appear less influenced by the frequency of losses, resulting in a more balanced distribution across the two advantageous options.

We compared the behaviors of humans and LLMs using the Prospect Valence Learning Model with a decay Reinforcement Learning rule~\cite{ahn2008comparison} (see \href{SI.com}{\textit{SI Appendix, Supplementary Note 2}} for details).
This model was often used to characterize decision-making processes in this task through four parameters: learning rate ($A$), choice consistency ($c$), outcome sensitivity ($\alpha$), and loss aversion ($\lambda$). Our analysis revealed significant differences in the posterior estimates of these parameters between LLMs and humans (Fig.~\ref{fig:IGT_fig1}C). Specifically, all the
LLMs showed a significantly higher learning rate ($A$), indicating a stronger reliance on cumulative past outcomes, which allowed them to identify and exploit patterns more effectively over time. In contrast, humans weighted recent and past outcomes more evenly, reflecting a more flexible but less pattern-oriented strategy.
 All the LLMs also exhibited higher choice consistency ($c$), meaning that they more reliably selected options with the highest learned expected value, with less variability or noise in their choices, whereas humans displayed lower consistency, indicating more stochastic or exploratory behavior.
\textcolor{black}{Additionally, LLMs generally showed higher outcome sensitivity ($\alpha$) and loss aversion ($\lambda$) compared to humans, with the exception of GPT, which did not significantly differ from humans in loss aversion. This suggests that LLMs typically tended to have a stronger reaction to outcomes, particularly negative ones, compared to humans.}

\subsection*{Decision-making under risk} 
The Cambridge Gambling Task assessed decision-making under risk by providing explicit information about the gains and losses associated with each choice, while the outcome of each choice remained probabilistic.
Players were shown a row of ten boxes, divided into two types (red and blue), with a hidden gold coin randomly assigned to one box in each round. Players predicted which box type contained the coin by placing a proportion of their bets. 
The ratio of red to blue boxes, which was explicitly shown, varied from a weak asymmetry (e.g., 6:4) to a strong asymmetry (e.g., 9:1), affecting risk levels. A stronger asymmetry increased the probability of the coin being in the majority type, thereby reducing risk.

\begin{figure*}[!t]
       \centering
       \includegraphics[width=0.9\linewidth]{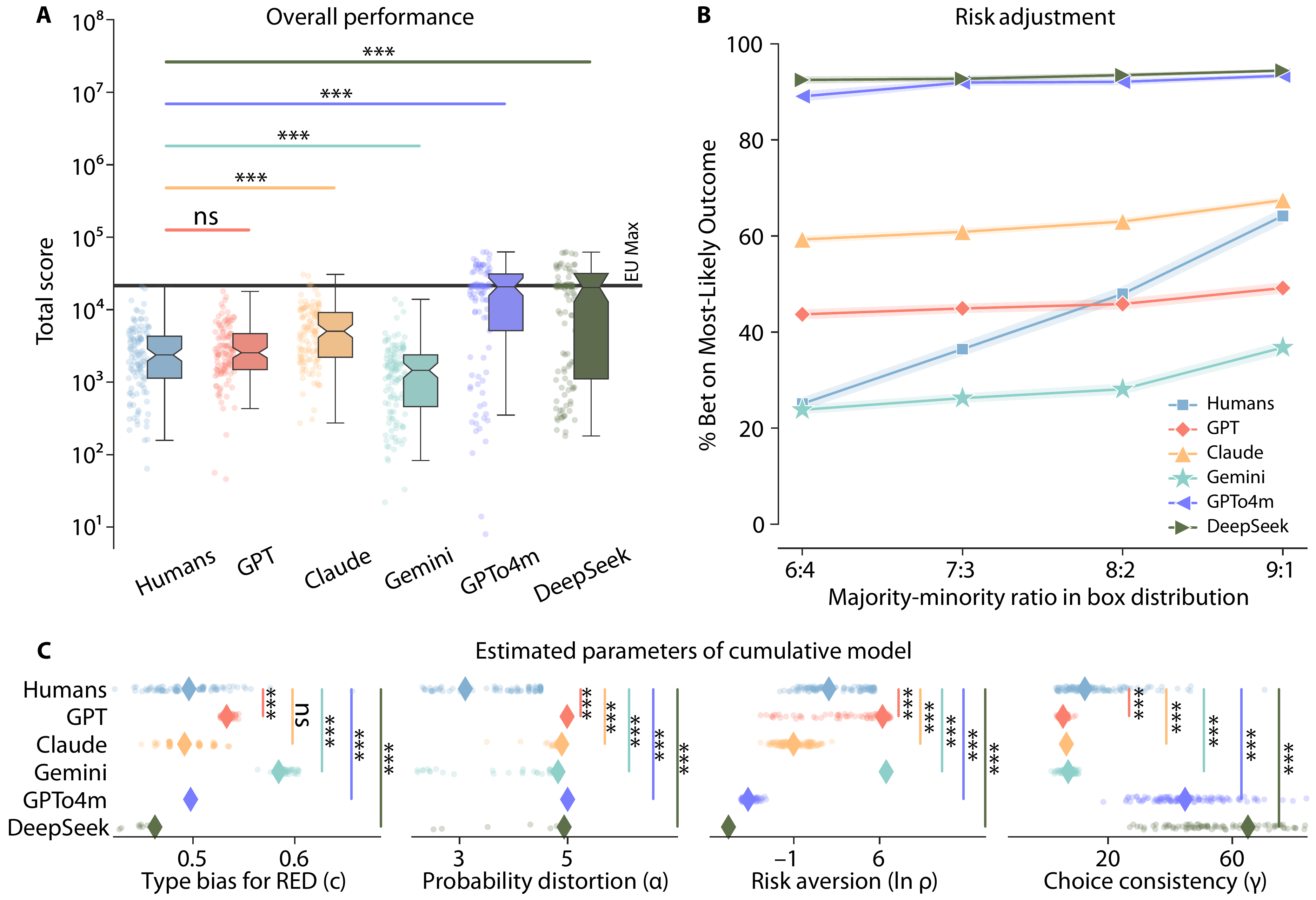}
       \vspace{1em}
       \caption{\textbf{LLMs, except for GPT and Gemini, outperformed humans on the Cambridge Gambling Task. However, all LLMs consistently exhibited a weaker tendency for risk adjustment and distinct parameter estimates in the cumulative model compared to humans.} \textbf{Panel (A)} shows the total scores of LLMs, human participants, and an Expected Utility Maximization (EU-max) strategy. A Mann-Whitney U test indicated that Claude, GPTo4m, and DeepSeek outperformed human participants, GPT matched human participants, while Gemini underperformed.
        \textbf{Panel (B)} illustrates the risk adjustment in terms of the proportions of bets placed on the most likely outcome. LLMs displayed a more stable and consistent betting behavior across varying levels of asymmetry in box distributions. In contrast, human participants raised their bets significantly as asymmetry intensified.
        \textbf{Panel (C)} presents the posterior estimates of four parameters in the cumulative model: type bias for RED boxes ($c$), probability distortion ($\alpha$), risk aversion ($\rho$), and choice consistency ($\gamma$). 
        A series of Mann–Whitney U tests revealed that, compared to human participants, GPT, GPTo4m, and Gemini exhibited significantly higher type bias for one choice ($c$),  while DeepSeek showed a significantly lower value in the opposite direction. Claude did not differ significantly from human participants on this parameter. 
        As for probability distortion ($\alpha$), all five LLMs displayed significantly greater distortion compared to humans. 
        Turning to risk aversion ($\rho$), GPT and Gemini exhibited significantly higher values than human participants, 
        while Claude, GPT4o4m, and DeepSeek showed significantly lower values. 
        Lastly, GPTo4m and DeepSeek exhibited significantly higher choice consistency, while GPT, Claude, Gemini, and humans showed relatively more variability. 
        All inferential (Mann–Whitney U tests) and descriptive statistics (means, medians, and standard deviations) for the Cambridge Gambling Task are reported in \href{SI.com}{\textit{SI Appendix}, Table~\ref{tab:cgt_results}}, including pairwise comparisons (panels a–b) and parameter estimates (panel c).
        All parameter estimates demonstrated satisfactory convergence with R-hat ($\hat{R}$) values below 1.01.}
       \label{fig:CGT_fig1}
\end{figure*}

We used total scores to assess the overall performance in this test.
As shown in Fig. \ref{fig:CGT_fig1}A, Claude, DeepSeek, and the two GPT models outperformed, or at least matched, human participants in total scores. In particular, both GPTo4m and DeepSeek achieved the highest total scores, with no statistically significant difference between them. Their performance was closer to that of a strategy optimized for expected utility maximization (see Methods for details). In contrast, Gemini performed worse than human participants, yielding the lowest overall score among the LLMs.
Following the convention~\cite{rogers1999dissociable}, we assessed decision-making quality by calculating the proportion of rounds in which players selected the majority  type, which reflected the ability to make probabilistically informed decisions.
As shown in \href{SI.com}{\textit{SI Appendix}, Fig.~\ref{fig:CGT_fig1_b}},  all the LLMs displayed near-ceiling decision-making quality, and consistently chose the majority type with proportions close to 1, regardless of the degree of asymmetry in box distributions.
In contrast, humans were unable to consistently choose the majority type, particularly in weak-asymmetry (e.g., a 6:4 red-to-blue ratio, representing high-risk) conditions.
These findings highlight the superior accuracy and consistency of LLMs in predicting the most likely outcome.  In contrast, humans showed greater variability and a higher tendency toward suboptimal choices, especially when the associated risk is high.

As shown in Fig.~\ref{fig:CGT_fig1}\textcolor{black}{B}, under high-risk conditions with weak asymmetry, humans adopted cautious strategies by placing a low proportion (e.g., 25\%) of their bets.
However, as asymmetry intensified (reducing risk), humans raised their bets significantly, aligning risk-taking with favorable odds of success to maximize rewards. 
Thus, humans demonstrated a dynamic and flexible approach to risk adjustment.
In contrast, LLMs displayed a more stable and consistent betting strategy across varying degrees of asymmetry, with less variation in risk adjustment.
Among the models, both GPTo4m and DeepSeek notably placed the highest bets (around 90\%) across all levels of the box distribution, indicating the strongest risk-taking tendency. Claude consistently placed high bets (above 60\%), indicating a propensity for risk-taking. GPT maintained moderate bet levels (generally below 50\%), suggesting a balanced but slightly risk-averse strategy. Gemini exhibited the most risk-averse behavior, with consistently low bets (typically under 30\%), but showed a noticeable increase in low-risk conditions. These results highlight a key distinction: humans flexibly adjusted their risk-taking in response to changing probabilities, while LLMs followed more rigid, consistent strategies regardless of risk level.

The Cumulative Model~\cite{romeu2020computational}  accounts for players’ decisions in this task using a probabilistic choice process (\href{SI.com}{\textit{SI Appendix, Supplementary Note 2}}).
This model incorporates four parameters: type bias for red ($c$), probability distortion ($\alpha$), risk aversion ($\rho$), and choice consistency ($\gamma$). 
The posterior estimates of these parameters showed significant differences between humans and LLMs (Fig. \ref{fig:CGT_fig1}C). \textcolor{black}{Human participants, Claude, and GPT4o4m showed minimal or no bias between response options, with type bias values close to 0.5. In contrast, Gemini exhibited the strongest bias toward red boxes, while DeepSeek showed a similarly strong bias in the opposite direction, with estimates substantially diverging from 0.5.} For probability distortion ($\alpha$), all the LLMs exhibited significantly higher levels of distortion than humans, indicating that LLMs tend to perceive high-probability events as even more likely and low-probability events as even less likely than they objectively are.
The risk aversion ($\rho$), however, differed across LLMs: GPT and Gemini showed greater risk aversion, whereas
Claude, GPTo4m, and DeepSeek showed strong risk-seeking tendencies. In contrast, human participants were more balanced---less risk-averse than GPT and Gemini, but also less risk-seeking than other LLMs.  Finally, choice consistency ($\gamma$) showed that GPTo4m and DeepSeek followed highly deterministic strategies, closely aligning with model-predicted expected values, while humans and other LLMs exhibited more variability in their choices.

\subsection*{Decision-making under set-shifting} 
The Wisconsin Card Sorting Task assessed decision-making with changing conditions. Participants were presented with items that vary in attributes such as color, shape, and the number of symbols. They were asked to match items to one of four cards based on a matching rule (i.e., an attribute), whereas the rule was not explicitly stated. Participants received feedback (`correct' or `incorrect') for their every match. Upon successfully achieving a predetermined number of correct matches, the rule would change without notice. Thus, this task required participants to identify the rule, detect the change, and continuously adapt to the new rule. 

\begin{figure*}[!t]
       \centering
       \includegraphics[width=\linewidth]{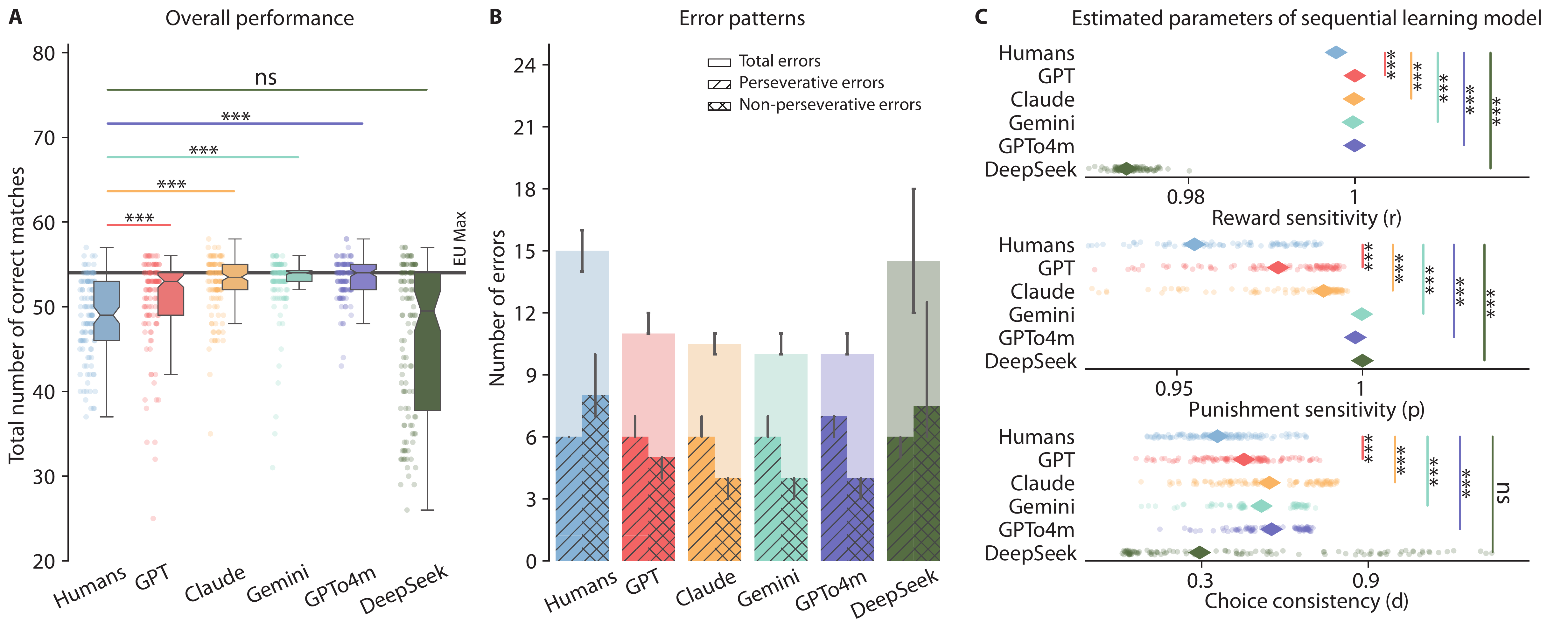}
       \vspace{1em}
       \caption{\textbf{All the LLMs outperformed, or at least matched, humans in the Wisconsin Card Sorting Task, while exhibiting generally distinct error patterns and parameter estimates in the sequential learning model compared to humans. }
        \textbf{Panel (A)} shows the total number of correct matches for human participants, LLMs, and an Expected Utility Maximization (EU-max) strategy. A two-tailed Mann-Whitney U test revealed that GPT, Claude, Gemini and GPTo4m all significantly outperformed human participants; however, DeepSeek matched human participants.
        \textbf{Panel (B)} illustrates comparisons of median across three error types made by human participants and LLMs. Bars represent the median number of total, perseverative, and non-perseverative errors, with 95\% confidence intervals of the median shown as error bars. Human participants made significantly more non-perseverative errors unrelated to the task than perseverative errors. In contrast, except for DeepSeek, LLMs produced more perseverative errors than non-perseverative ones.  
        The error pattern of DeepSeek matched that of human participants, with more non-perseverative errors than perseverative ones.
        \textbf{Panel (C)} presents the posterior estimates of the parameters in the sequential learning model. Mann-Whitney U tests revealed that compared to human participants,
        LLMs except for
        DeepSeek demonstrated higher 
        reward sensitivity ($r$), 
        greater sensitivity to punishment ($p$), and 
        higher choice consistency ($d$). As an exception, compared to humans, DeepSeek showed lower reward sensitivity, higher punishment sensitivity, and a comparable level of choice consistency. 
        All inferential (Mann–Whitney U tests) and descriptive statistics (means, medians, and standard deviations) for the Wisconsin Card Sorting Task are reported in \href{SI.com}{\textit{SI Appendix}, Table~\ref{tab:wcst_results}}, including pairwise comparisons (panels a–c) and parameter estimates (panel d).
        All parameter estimates demonstrated satisfactory convergence with R-hat ($\hat{R}$) values below 1.01.}
    \label{fig:WCST_fig1.pdf}
\end{figure*}

To assess performance, we examined the total number of correct matches. As shown in Fig. \ref{fig:WCST_fig1.pdf}A, all the LLMs, except for DeepSeek, significantly outperformed human participants, and approached the performance of a strategy optimized for expected utility maximization (see Methods for details). Specifically, Gemini and GPTo4m achieved the highest median number of correct matches. \textcolor{black}{Gemini exhibited the lowest variance, followed closely by GPTo4m, indicating that both models performed accurately and consistently.}

We further analyzed their decision-making using four commonly used metrics~\cite{glascher2019model}: 
the number of rounds to complete the first set (TRSET1), failure to maintain set (FSET), perseverative errors, and non-perseverative errors.
As indicated by TRSET1, compared to humans, \textcolor{black}{all the LLMs, except for DeepSeek, used fewer rounds to complete the first set, suggesting that almost all LLMs more efficiently identified and applied the underlying rule (\href{SI.com}{\textit{SI Appendix}, Fig.~\ref{fig:WCST_fig2_TRSET1_FSET})}.}
The FSET measures how often participants changed their matching strategy after achieving five or more consecutive correct matches without negative feedback (`incorrect'). Humans and LLMs displayed comparable values of FSET, suggesting that they were equally capable of maintaining a correct strategy once identified (\href{SI.com}{\textit{SI Appendix},  Fig.~\ref{fig:WCST_fig2_TRSET1_FSET}}).
The perseverative errors are the number of incorrect matches caused by applying a previously learned rule to a newly changed condition. In contrast, the non-perseverative errors are random errors unrelated to the task. As shown in Fig. \ref{fig:WCST_fig1.pdf}B,  human participants tended to make more non-perseverative errors than perseverative ones. However, \textcolor{black}{LLMs, except for DeepSeek, exhibited the opposite pattern and produced more perseverative errors than non-perseverative ones. DeepSeek, in contrast, matched the pattern of humans by showing more non-perseverative errors.}

The Sequential Learning Model \cite{bishara2010sequential} describes how participants adjust their decision in the Wisconsin Card Sorting Task through three parameters: reward sensitivity ($r$), punishment sensitivity ($p$), and choice consistency ($d$) (\href{SI.com}{\textit{SI Appendix, Supplementary Note 2}}). As shown in Fig.~\ref{fig:WCST_fig1.pdf}C, the posterior estimates for these parameters revealed significant differences between LLMs and humans. For both reward sensitivity ($r$) and punishment sensitivity ($p$), LLMs typically exhibited higher values than humans, suggesting that they adapted more quickly to both positive and negative feedback. 
Choice consistency ($d$) showed that LLMs generally made more deterministic decisions consistent with model-predicted expected values, in contrast to the greater variability observed in human choices. The only exception was DeepSeek, which did not differ significantly from humans in choice consistency and was less sensitive to rewards compared to humans. However, it was more sensitive to punishment than humans.
 
\section*{Discussion}
By virtue of passing the Turing test, LLMs have to possess decision-making capabilities, but most of the actual studies about them have been restricted to domain-specific settings, where multiple decision dimensions are often intertwined.  Focusing on decision-making under uncertainty, risk, and set-shifting, we used three standard psychological tests to investigate five LLMs. Notably, four of them---GPT-4o, Claude, GPTo4m, and DeepSeek---consistently matched or exceeded human performance across all tasks. Moreover, their performance generally approached optimality. Our findings offer evidence for the general decision-making abilities of LLMs, particularly in navigating uncertainty, calibrating risk, and adapting flexibly to changing conditions.

The AI's super-human performance also comes with an apparent non-human cognition. LLMs often employed strategies that diverged substantially from those of human participants. 
In the Iowa Gambling Task, LLMs were more sensitive to the frequency of losses, leading to divergent preferences between the two advantageous decks. In contrast, human choices were less influenced by loss frequency and showed a more balanced distribution across those decks. 
In the Cambridge Gambling Task, LLMs made optimal probabilistic choices by consistently selecting the option with the highest expected return, yet showed minimal adjustment in their betting strategies in response to changing levels of risk. In contrast, human participants exhibited greater variability in their betting behavior despite making less optimal probabilistic choices. In the Wisconsin Card Sorting Task, LLMs adapted to rule changes faster and made fewer random errors, while humans were more prone to make mistakes. Taken together, these findings suggest that LLMs act as if relying on cognitive processes that diverge substantially from those of humans. Therefore, replacing humans with machines in decision-making at a large scale will have profound consequences for our society.

Our results underscore the risks associated with using LLMs as substitutes for human participants in behavioral research and practical decision-making.
The models not only fail to reproduce typical human decision-making but also can generate misleading signals. A notable example is their consistent lack of risk adjustment, a trait that most humans exhibit. This deficiency is not primarily a deviation from normative behavior; rather, it is a sensitive trans-diagnostic marker of cognitive dysfunction, widely seen across a range of psychiatric conditions such as addiction and depression~\cite{clark2008differential,effah2024exploring}. 
As such, relying on LLMs in tasks involving probabilistic reasoning can obscure important psychological insights.
Moreover, LLMs also do not capture the breadth of reasoning strategies typical of human cognition. Their generated responses are less diverse than those of human participants and remain largely unaffected by demographic cues or contextual framing, according to our robustness checks (see Methods, and \href{SI.com}{SI Appendix, Supplementary Note 3}). In contrast, human decisions are shaped by individual differences and context, such as age, gender, and cultural background~\cite{byrnes1999gender,weber1998cross,tversky1981framing,tymula2013like}. Thus, using LLMs to replace human participants in behavioral research risks oversimplifying the phenomena under study and misrepresenting how people respond in actual decision-making contexts.

If LLMs do not faithfully replicate human cognition, what exactly are they emulating? 
Our parameter estimation from computational models suggests that LLMs excel at recognizing and exploiting historical patterns, enabling them to efficiently gather information under uncertainty. They also exhibit systematic probability distortions, allowing their choices under risk to align closely with the options most likely to succeed. Across tasks, LLMs tend to show heightened sensitivity to outcomes, particularly negative ones, and generally make decisions consistent with the principle of expected utility, reflecting a form of rationality. These behavioral tendencies appear to underpin their strong performance in decision-making tasks.
 Moreover, we found that LLMs' decision-making patterns remained largely consistent across different prompt formulations and temperature settings (\href{SI.com}{\textit{SI Appendix}, Supplementary Note 3}). 
These findings suggest that while LLMs do not behave as cognitive replicas of human decision-making, they operate as consistent, rational, and outcome-driven agents. 

Despite differences in risk attitudes (e.g., Claude being more risk-seeking, while GPT and Gemini are more risk-averse), their behavior is primarily shaped by the structural features of the tasks they encounter.

While a consistent, rational, outcome-driven agent may be appealing from a utilitarian perspective, it does not necessarily translate into human acceptance of AI. In our post-experiment survey (\href{SI.com}{\textit{SI Appendix},  Fig.~\ref{fig:ai_attitude}}), participants expressed little interest in seeking assistance from AI, even when they were informed of the LLMs’ high performance on the same tasks and despite their own lower scores. Furthermore, participants did not believe that AI could improve their decision-making, either in terms of performance or efficiency. These findings point to a form of algorithm aversion~\cite{dietvorst2015algorithm,castelo2019task,karatacs2023thinking}, which may arise from concerns about trust, autonomy, or the perceived appropriateness of algorithmic input, particularly in decisions involving ambiguous outcomes, complex judgment, or personal values. Thus, even when LLMs demonstrate high technical performance, their effective integration into human decision-making processes will require attention not only to their capabilities but also to users’ perceptions, interpretations, and willingness to accept algorithmic guidance.

Even if people were willing to adopt LLMs to assist in their decisions, it is important to ask: when and for what kinds of decisions can these systems provide meaningful assistance?  LLMs tend to perform well in settings with clearly defined goals and outcomes, thanks to their rationality and outcome sensitivity. Yet humans, famously sub-optimal decision makers, often succeed by departing from strict rationality and discounting cost-benefit logic.
Consider the Wright Brothers: their crash-prone test flights, many ending in failure, defied utility maximization principles but drastically accelerated the path to powered flight. In such cases, heuristics, imagination, and a willingness to tolerate error or loss can outperform conservative or algorithmically optimized strategies. LLMs, bound to rational evaluation functions, may struggle to replicate this productive `irrationality,' which remains a distinctly human advantage.

Overall, decision-making is never solely about performance. Many human choices cannot be objectively evaluated, and what seems like a poor decision now may later prove wise. In such cases, what truly matters is how the decision is made---how we reason, weigh uncertainty, balance risks, and adapt to change.
While our findings show that LLMs can perform well on stylized psychological decision tasks, they also reveal a crucial disconnection between human-level performance and human-like cognition. 
This suggests that LLM intelligence should not be viewed as a mirror of human thought, but rather as a novel form of reasoning.
For system designers, the issue is not just performance, but whether people are willing to rely on LLMs for decision-making, especially when LLMs do not think or act like them.  For policymakers, the challenge goes beyond evaluating performance: it requires assessing whether, and to what extent, these systems reason in ways that align with human values and norms, and uphold accountability. For everyday users, the question is how much autonomy we are willing to share with these systems, especially in decisions with real consequences. After all, a decision is also about what kind of reasoning we are willing to trust, and the kind of minds we are willing to build. These decisions not only reflect who we are but also shape who we may become, and at times, they may alter the course of history.

\subsection*{Methods}

\paragraph{Iowa Gambling Task}
This task tests if participants can prioritize long-term benefits over short-term gains in an uncertain setting~\cite{bechara1994insensitivity}. 
In our experiments, this task lasted for 80 rounds and comprised four decks (labeled as A, B, C, and D), from which participants must choose one deck per round. 
Each choice offered an immediate reward, with the possibility of an occasional penalty. Decks A and B were disadvantageous decks, while C and D were advantageous decks.
Specifically, each choice of A or B resulted in a high immediate reward, but the penalty, when it occurred, was substantial, offsetting the immediate rewards in the long run. As a result, the expected payoffs from these decks were lower.
Conversely, each choice of C or D resulted in a low immediate reward, though the penalty, when it occurred, was minor, yielding a higher expected payoff in the long run.
Additionally, compared to Deck A (or C), 
the penalty from Deck B (or D) occurred less frequently. 
Participants were not told the total number of rounds or the reward and penalty associated with each deck.
Their decisions were made under uncertainty and relied on learning over time. This task can be mathematically framed as a multi-armed bandit problem, for which the upper confidence bound~\cite{auer2002finite} and $\epsilon$-greedy~\cite{watkins1989learning} are two well-established solution strategies.

\paragraph{Cambridge Gambling Task}
This task  tests decision-making under risk~\cite{rogers1999dissociable}. In our experiments, this task consisted of 64 rounds in total, with every 8 rounds forming a sub-session. Participants saw a row of 10 boxes categorized into two types: Red and Blue; they must choose under which box type (Red or Blue) a gold coin was hidden and must place a bet on the type chosen. The ratio of red to blue boxes was explicitly shown and indicated the risk associated with each choice. 
There were 8 possible red-to-blue ratios: 1:9, 2:8, 3:7, 4:6, 6:4, 7:3, 8:2, and 9:1.
A stronger asymmetry (e.g., 1:9) in the red-to-blue ratio indicated a higher probability of the coin being in the majority type, suggesting a lower risk in choosing the majority type.
Bet levels were fixed at 5\%, 25\%, 50\%, 75\%, or 95\% of their current score. A correct guess resulted in gaining scores equal to the bet, while an incorrect guess resulted in losing scores equal to the bet.
Different from the original design, we presented all five bet options simultaneously, rather than sequentially, as the sequential presentation aimed to test human impulsive behaviors, which were not applicable to LLMs.
Participants had complete information about the number of rounds, the risk (the red-to-blue ratios), and the possible rewards and penalties associated with each choice.  The expected utility maximization strategy in this context is to always choose the majority box type and place the highest possible (95\%) bet.  However, this strategy does not reflect typical decision-making behavior in healthy individuals, as it lacks risk adjustment~\cite{clark2008differential}. 

\paragraph{Wisconsin Card Sorting Task}
This task tests participants’ ability to make decisions under set-shifting conditions~\cite{berg1948simple}. 
In our experiments, this task lasted 64 rounds, and comprised four cards (labeled as A, B, C, and D). In each round, participants were presented with an item and required to choose one card that matched the item's pattern. Both the cards and items were marked with a set of symbols, and the matching pattern was based on one of three attributes of the symbols: color, shape, or number. A correct match resulted in a reward, while an incorrect match resulted in no points.
The matching rule remained consistent for a sequence of eight consecutive correct responses, after which it changed without explicit notification. Participants were informed that the matching rule could change but were not told when or how. This design required participants to adapt their decision-making strategy in response to shifting conditions over time.
The expected utility maximization strategy for this task is to switch to a different attribute after receiving negative feedback and to continue using the current attribute following positive feedback.

\paragraph{Large Language Models}  
We provided LLMs with a system prompt and a decision-making prompt (see \href{SI.com}{\textit{SI Appendix}, Supplementary Note 1} for prompt details). Through the system prompt, LLMs received exactly the same experimental instructions as human participants. 
The decision-making prompt presented LLMs with their previous choices and the corresponding outcomes from past rounds, and asked LLMs to make a choice for the current round. However, just like human participants, LLMs were not given explicit instructions on how to utilize the provided information.
We maintained the default parameters for all LLMs unless specified otherwise, as this better simulates typical user conditions, which served as the baseline.
To prevent LLMs from exploiting biases to solve the tasks~\cite{zhao2021calibrate}, we performed a cyclic permutation of the order of options for all tasks. 

\paragraph{Robustness Checks}
To assess if our findings on LLMs are robust to prompt variations, we conducted robustness checks by varying the experimental settings as follows (see \href{SI.com}{\textit{SI Appendix}, Supplementary Note 3} for details):  
 i) adjusting the LLMs' temperature from the default value (i.e.\ 1) to 0 and 0.5, 
 ii) applying arithmetic transformation to the scores of each round (e.g., multiplying them by a constant factor or adding a constant value),
 iii) restructuring the prompts to reflect various decision-making contexts (including both economic and medical scenarios), and  
 iv) introducing role-play prompts to test the impact of demographic information (e.g., different age ranges, genders, and ethnicity) and risk preferences (e.g., risk-taking and risk-averse).
 In total, there were 19 variants in the Iowa Gambling Task as well as the Cambridge Gambling Task, and 15 variants in the Wisconsin Card Sorting Task. We repeated each variant 10 times using GPT-4o.
Overall, we observed that the decision-making patterns of LLMs remained qualitatively unchanged across these variants ( \href{SI.com}{\textit{SI Appendix},  Fig.~\ref{fig:igt_gpt_4o_robust}, ~\ref{fig:cgt_gpt_4o_robust_qm}, ~\ref{fig:cgt_gpt_4o_robust_ra}, ~\ref{fig:wcst_gpt_4o_robust}}).

 \paragraph{Human Subjects}
We recruited a total of 360 participants, including 37.8\% women, with a mean age of 18.95 years (Table \ref{table:human_information}). The experiments were pre-registered (AsPredicted \#182473, \#186129, and \#203115) and were conducted at Northwest A\&F University and Northwestern Polytechnical University in Xi’an, from July to December 2024. Participants came from multiple departments to minimize interaction among them. Each task was administered using oTree~\cite{chen2016otree} and completed by 120 participants. 
Details of these tasks were maintained under strict confidentiality until participants arrived at the lab. Upon arrival, each participant was allocated a computer, isolated from others by partitions to guarantee independent completion. 
Participants began by a tutorial that described the task (\href{SI.com}{\textit{SI Appendix}, Fig.~\ref{fig:intro_of_igt}, \ref{fig:intro_of_cgt}, \ref{fig:intro_of_wcst}}). Then, the experiment, which lasted for multiple rounds, started. Each round comprised a choice page, where participants made a decision in a task,  and a result page, where the score of each round and the cumulated score were shown (\href{SI.com}{\textit{SI Appendix}, Fig.~\ref{fig:choice_result_of_igt}, \ref{fig:choice_result_of_cgt}, \ref{fig:choice_of_wcst}, \ref{fig:result_of_wcst}}). 
After completing all rounds, participants saw their final scores (\href{SI.com}{\textit{SI Appendix}, Fig.~\ref{fig:final_result}}), and then their demographic data (\href{SI.com}{\textit{SI Appendix}, Fig.~\ref{fig:demographic}}), feedback on the task, and attitudes toward artificial intelligence (\href{SI.com}{\textit{SI Appendix}, Fig.~\ref{fig:question}}) were collected.
Throughout the experiment, communication between participants was strictly prohibited. Each session lasted approximately 30 minutes. We provided an average payment of 30 CNY, which included a 15 CNY show-up fee, with the remaining amount being calculated based on experiment scores.

\paragraph{Ethics Statement}

This study was approved by the Northwestern Polytechnical University Ethics Committee on the use of human participants in research and carried out in accordance with all relevant guidelines. Informed consent was obtained from all participants.

\paragraph{Data, Materials, and Software Availability.}  
Data, Materials, and Software Availability. Data and code for the current study are available through the \href{https://github.com/ynulihao/LLM_vs_Human_Decision_Making}{GitHub repository}.

\newpage
\bibliography{sn-bibliography}
\newpage

\appendix 
\begin{appendices}
\setcounter{figure}{0}
\renewcommand{\thefigure}{S\arabic{figure}}
\setcounter{table}{0}
\renewcommand{\thetable}{S\arabic{table}}

\section*{Supplementary note 1: Prompt for LLMs}
The prompts provided to the LLMs consist of two parts: the system prompt and the decision-making prompt. Through the system prompt, LLMs received exactly the same experimental instructions as human subjects, with no additional information about tasks. The decision-making prompt presents LLMs with their previous choices and the corresponding outcomes from past rounds, and asks LLMs to make a choice for the current round. However, just like participants, LLMs are not given explicit instructions on how to utilize the provided information.

\subsection*{Iowa Gambling Task}
\begin{figure}
  \centering
  \includegraphics[width=\textwidth]{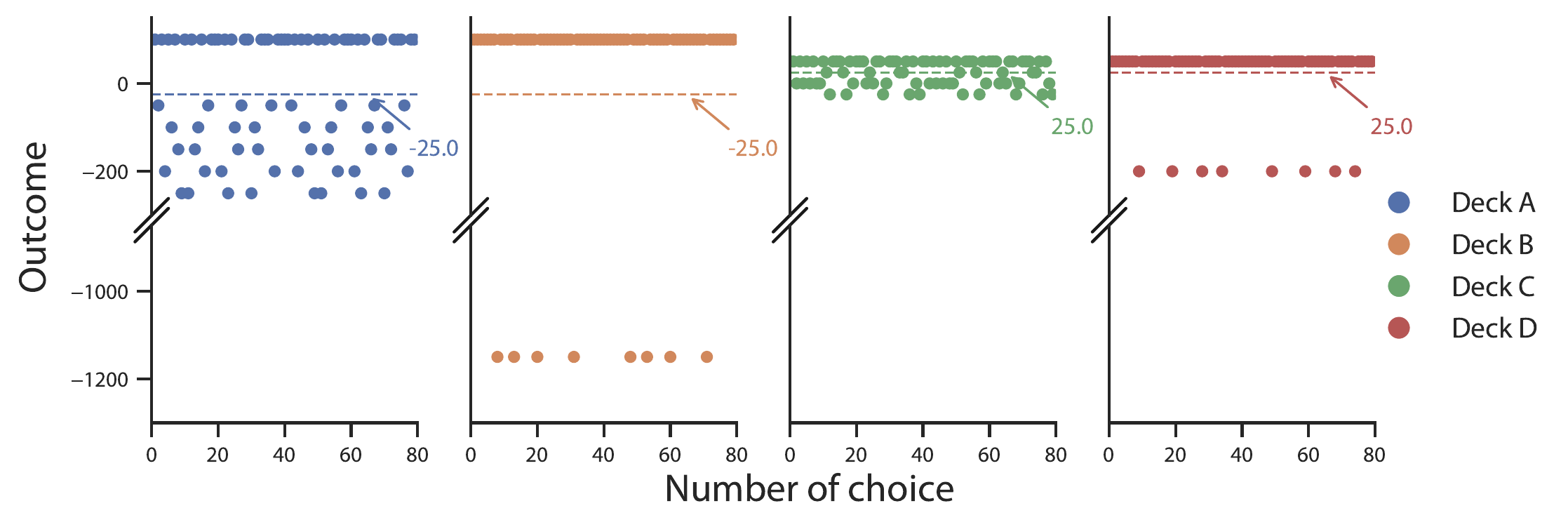}
  \vspace{1em}
  \caption{The outcomes for four decks (A, B, C, D) based on the number of choices, with dashed lines representing the average outcome for each deck. Decks A and B provide higher immediate rewards but result in a net loss for every 10 choices due to penalties. Deck A offers five small penalties, while Deck B offers one large penalty. In contrast, Decks C and D provide lower immediate rewards but yield a net gain for every 10 choices. Deck C offers five small penalties and Deck D offers one large penalty.}
  \label{fig:igt_outcome}
\end{figure}

Fig.~\ref{fig:igt_outcome} shows the outcomes for each choice of these decks. 
\verb|<choice_{i}>|, \verb|<reward_{i}>|, and \verb|<penalty_{i}>| represent the choice made, the reward received, and the penalty incurred (if any) in the \(i\)-th round, respectively. %
\verb|<points_so_far>| indicates the cumulative payoff accumulated by the LLM up to the current round.

\begin{promptbox}[System prompt]
  \begin{verbatim}
  In this game, you find yourself in a mysterious room with four ancient treasure chests. Opening each chest will yield a reward but may also simultaneously result in a penalty, depending on the chosen chest. With each turn, you will choose one chest to open. Please consider carefully, as your choice may significantly impact your points. Specifically, the rewards will increase your points, while penalties will deduct your points. At the start of the game, you will receive a loan of 2000 points. The game has several rounds in which your points will accumulate, and your goal is to maximize your points by the end of the game.

  The only hint I can give you, and the most important thing to note is this: Out of these chests, there are some that are worse than others, and to win you should try to stay away from bad chests. No matter how much you find yourself losing, you can still win the game if you avoid the worst chests. Also note that the computer does not change the order of the chests once the game begins. It does not make you lose at random, or make you lose money based on the last chest you picked.
  
  Your response must always present in the following format:

  <reasoning>Reasons for your choice this round</reasoning>
   
  <choice>Any number between 1-4 indicates your choice of chest</choice>
  \end{verbatim}
\end{promptbox}

\begin{promptbox}[Decision-making prompt]
  \begin{verbatim}
  Here is the historical information from the past round(s), and you may use it as a reference for your following choice.

  In round 1, you chose chest number <choice_{1}>. You earned <reward_{1}>  points in rewards (and received a penalty of <penalty_{1}> points).
  In round 2, you chose chest number <choice_{2}>. You earned <reward_{2}>  points in rewards (and received a penalty of <penalty_{2}> points).
  \end{verbatim}
  \[
      \vdots
  \]
  \begin{verbatim}
  In round i-1, you chose chest number <choice_{i-1}>. You earned <reward_{i-1}>  points in rewards (and received a penalty of <penalty_{i-1}> points). 

  Your total points so far: <points_so_far> points. Now this is the i-th round of the game. Please make your choice. 
  \end{verbatim}
\end{promptbox}

\subsection*{Cambridge Gambling Task}

\verb|<round_interactions>| represents the number of rounds in each subsession. 
\verb|<choice_type_{i}>| represents the type of the box chosen in the \(i\)-th round, and \verb|<choice_percent_{i}>| represents the bet proportion in the \(i\)-th round. %
\verb|<coin_type_{i}>| represents the type of the box where the coin is placed. %
\verb|<payoff_{i}>| represents the reward (or penalty) received in the \(i\)-th round. %
\verb|<remain_by_subsession>| represents the player's total earnings in the current subsession. %
\verb|<num_type_RED_boxes>| and \verb|<num_type_BLUE_boxes>| represent the number of Type RED and Type BLUE boxes in the current round.

\begin{promptbox}[System prompt]
  \begin{verbatim}
    In front of you are 10 treasure chests, some of which are type F and others are type J. A gold coin is hidden in one of these chests. Your task is to guess in which type of chest the gold coin is hidden.
    Every <round_interactions> rounds constitute a phase. At the start of each phase, your points will be set to 100. Your goal is to maximize the points in each phase, and the sum of the points from each phase is your total score in the game.
    
    Each round, you need to bet by deducting a certain percentage from your existing points to choose the type of chest. The available percentages are: 5%, 25%, 50%, 75%, 95%. If the gold coin is hidden in any one of the chests of the type you guessed, you win; otherwise, you lose. If you win, you will receive double the points of your bet; if you lose, you gain nothing. Points are rounded to the nearest whole number.
    An important game setting is: the position of the gold coin is completely random in each round and has no connection to previous locations. That is, each round's coin position is an independent event.

    Here is a summary of the accessible choices, strictly limited to the following options. Each choice maps to a specific type guess and bet percentage, with the list of choices and their corresponding as follows:

    Choice 0 maps to F, 5% bet.
    Choice 1 maps to F, 25% bet.
    Choice 2 maps to F, 50% bet.
    Choice 3 maps to F, 75% bet.
    Choice 4 maps to F, 95% bet.
    Choice 5 maps to J, 5% bet.
    Choice 6 maps to J, 25% bet.
    Choice 7 maps to J, 50% bet.
    Choice 8 maps to J, 75% bet.
    Choice 9 maps to J, 95% bet.

    Your response must always present in the following XML format:
    <reasoning>Reasons for your choice this round should cover why you chose the type and bet percentage</reasoning>
    <choice>Any number between 0-9 indicates your choice</choice>
    The available range for <choice> is 0-9.
  \end{verbatim}

\end{promptbox}

\begin{promptbox}[Decision-making prompt]
  \begin{verbatim}
    Here is the historical information from the past round(s), and you may use it as a reference for your following choice.

    In round 1, you chose the <choice_type_{1}> chest and bet <choice_percent_{1}>%. (Un)Fortunately, the coin was hidden under the <coin_type_{1}> chest, and you received <payoff_{1}> points as a reward (or penalty).
    In round 2, you chose the <choice_type_{2}> chest and bet <choice_percent_{2}>%. (Un)Fortunately, the coin was hidden under the <coin_type_{2}> chest, and you received <payoff_{2}> points as a reward (or penalty).
  \end{verbatim}

  \[
    \vdots
  \]

  \begin{verbatim}
    In round i-1, you chose the <choice_type_{i-1}> chest and bet <choice_percent_{i-1}>%. (Un)Fortunately, the coin was hidden under the <coin_type_{i-1}> chest, and you received <payoff_{i-1}> points as a reward (or penalty).

    Your total points in this phase so far: <remain_by_subsession> points. Now this is the i-th round of the game. In front of you are <num_type_RED_boxes> Type F chest(s) and <num_type_BLUE_boxes> Type J chest(s). Please make your choice.
  \end{verbatim}

\end{promptbox}

\subsection*{Wisconsin Card Sorting Task}
\verb|<chest_attribute_{A}>| describes the symbols on Card~A (e.g., ``Card A has 2 green flowers.''). %
\verb|<item_attribute_{i}>| describes the characteristics of the item in the \(i\)-th round (e.g., ``The item has 2 blue hearts.''). %
\verb|<choice_id_{i}>| indicates the card selected in the \(i\)-th round (A, B, C, or D). %
\verb|<reason_{i}>| explains the reasoning process for the \(i\)-th round, which aims to improve the LLM's performance on this task. 

\begin{promptbox}[System prompt]
  \begin{verbatim}

    In the game, you have 4 chests in front of you.
    In each round, you will be presented with one item, and your task is to choose one of the 4 chests to match the presented item based on its pattern.
    The pattern will be one of the following three: color, shape, or number. There will be no combination of these patterns to define the match.
    If the match is correct, you will receive a ``Match Correct''; if incorrect, you will get a ``Match Failed.''
    Note: You must determine whether to match based on color, number, or shape. Once you figure out the rule, you can follow it for a while, but stay alert—the rule changes periodically! Pay close attention to feedback; if you receive error messages, it's time to adjust your rule. That's all!

    <chest_attribute_{A}>.

    <chest_attribute_{B}>.

    <chest_attribute_{C}>.

    <chest_attribute_{D}>.

    Your response must always present in the following format:
    <reasoning>A brief reason for your choice this round</reasoning>
    <choice>Any number between 1-4 indicates your choice of chest A, B, C, D</choice>
  \end{verbatim}

\end{promptbox}

\begin{promptbox}[Decision-making prompt]
  \begin{verbatim}
    Here is the historical information from the past round(s), and you may use it as a reference for your following choice.

    In round 1, <item_attribute_{1}>, You chose chest <choice_id_{1}>. Your reasoning process is <reason_{1}>. Match Correct (or Failed).
    
    In round 2, <item_attribute_{2}>, You chose chest <choice_id_{2}>. Your reasoning process is <reason_{2}>. Match Correct (or Failed).
    
  \end{verbatim}

  \[
    \vdots
  \]

  \begin{verbatim}

    In round {i-1}, <item_attribute_{i-1}>, You chose chest <choice_id_{i-1}>. Your reasoning process is <reason_{i-1}>. Match Correct (or Failed).
    
    Now this is the i-th round of the game. <item_attribute_{i}>. Please make your choice.
  \end{verbatim}

\end{promptbox}

\section*{Supplementary note 2: Computational models}
We used computational models to quantitatively understand the  decision-making processes of  humans and LLMs.  
We utilized
hierarchical Bayesian modeling to estimate the parameters of these models, employing the Stan software package~\cite{carpenter2017stan} for posterior inference. The implementation refers to the hBayesDM package~\cite{hBayesDM}. The descriptions of each model are as follows.

\subsection*{Iowa Gambling Task}
We used the Prospect Valence Learning Model with decay Reinforcement Learning rule~\cite{ahn2008comparison} in the Iowa Gambling Task, which includes four parameters: learning rate ($A$),  choice consistency ($c$), outcome sensitivity ($\alpha$), and loss aversion ($\lambda$).
This model employs the prospect theory utility function to evaluate outcomes, which is characterized by a diminishing sensitivity to both gains and losses (parameterized by $\alpha$) and an asymmetric weighting of losses relative to gains (parameterized by $\lambda$). The utility function $u(t)$ for outcome $x(t)$ at the round $t$ is defined as

$$
u(t) = \begin{cases}
x(t)^\alpha & \text{if } x(t) \geq 0, \\
-\lambda |x(t)|^\alpha & \text{if } x(t) < 0.
\end{cases}
$$
Let $E_j(t)$ represent the expected value of the $j$-th deck at the round $t$, and $A$ represent the learning rate governing the forgetting of historical information. The indicator $\delta_j(t)$ denotes whether the $j$-th deck was selected at round $t$, thereby influencing the update of its expectancy
$$
E_j(t) = A \cdot E_j(t-1) + \delta_j(t) \cdot u(t).
$$
The model utilizes the softmax rule to select a deck for the round $t+1$, i.e.
$$
\operatorname{Pr}(j,t) = \frac{e^{\theta \cdot E_j(t)}}{\sum_{k=1}^4 e^{\theta \cdot E_k(t)}},
$$
where $\theta$ is conventionally set to $3^c - 1$~\cite{ahn2008comparison}, with $c$ being the choice consistency parameter. 

\subsection*{Cambridge Gambling Task} 
We used the Cumulative Model~\cite{romeu2020computational} in the Cambridge Gambling Task, which involves four parameters: type bias for RED ($c$), probability distortion ($\alpha$), risk aversion ($\rho$), and choice consistency ($\gamma$). 
The probabilities of selecting red or blue boxes in this model are defined as follows:
$$
\operatorname{Pr}(Red) = \frac{c r^\alpha}{c r^\alpha + (1-c)(1-r)^\alpha}, \quad \operatorname{Pr}(Blue) = 1 - \operatorname{Pr}(Red),
$$
where $r$ represents the proportion of red boxes, $c$ represents the type bias for selecting red boxes, and $\alpha$ is the probability distortion parameter, indicating whether the participant distorts the asymmetry in probabilities.
After selecting a specific box type, for each betting proportion $b_i \in \{5\%, 25\%, 50\%, 75\%, 95\% \}$, the expected value is given as follows:
\[
\begin{aligned}
E(b_i \mid \text{Chosen type}) &= \operatorname{Pr}(\text{Chosen type}) \cdot u(\text{Wins}) \\
&\quad + \big(1 - \operatorname{Pr}(\text{Chosen type})\big) \cdot u(\text{Loss}),
\end{aligned}
\]
where $ u(\text{Wins}) $ and $ u(\text{Loss}) $ correspond to the utility functions under the winning and losing outcomes, respectively. These two utility functions are defined as follows: 
\[
u(\text{Wins}) = \log\big(1 + \text{CurrScore} \cdot (1 + b_i)\big)
\]
\[
u(\text{Loss}) = \log\big(1 + \rho \cdot \text{CurrScore} \cdot (1 - b_i)\big),
\]
where $ \text{CurrScore}$ represents the current score in the current sub-session, and $\rho$ is the risk aversion parameter.
The softmax rule is applied to determine the betting proportion, i.e. 
\[
\operatorname{Pr}\left(b_i \mid \text{Chosen Type}\right) = 
\frac{\exp \big\{\gamma \cdot E\big(b_i \mid \text{Chosen Type}\big)\big\}}
{\sum_j \exp \big\{\gamma \cdot E\big(b_j \mid \text{Chosen Type}\big)\big\}}.
\]
Note that we reworded the experiments to mitigate the memorization effect. After rewording, types F and J correspond to red and blue boxes, respectively.

\subsection*{Wisconsin Card Sorting Task} 
We applied the Sequential Learning Model~\cite{bishara2010sequential} in the Wisconsin Card Sorting Task, which incorporates three parameters: reward sensitivity ($r$), punishment sensitivity ($p$), and choice consistency ($d$).
For each presented card $k$,
let $\mathbf{m}_{k}(t)$ be a $3 \times 1$ vector, in which each generic element $m_{k,i}(t) = 0 \text{ or } 1$ denotes  whether the card $k$ matches the item at  round $t$ according to a specific rule $i \in \{\text{color}, \text{shape}, \text{number} \}$. 
Let $\mathbf{a}(t)$ be a $3 \times 1$ attention vector representing the attentional weights assigned to each rule at the round $t$.
The probability of selecting card $k$ at the round $t$ is given by $\mathrm{Pr}(k,t)$:
$$
\operatorname{Pr}(k,t)=\frac{\mathbf{m}_{k}(t)^{\top} \mathbf{a}(t)^d}{\sum_{j=1}^4 \mathbf{m}_{j}(t)^{\top}\mathbf{a}(t)^d},
$$
where \(d\) represents the choice consistency parameter, and $\mathbf{a}(t)^d$ denotes the element-wise exponentiation of the vector 
$\mathbf{a}(t)$. 
After selecting card $k$, the attention vector $\mathbf{a}({t+1})$ for the round $t+1$ is updated based on the feedback signal, such that 
\[
\mathbf{a}({t+1}) = 
\begin{cases}
(1 - r)\mathbf{a}(t) + r\mathbf{s}(t) & \text{if `correct'} \\
(1 - p)\mathbf{a}(t) + p\mathbf{s}(t) & \text{if `incorrect'}
\end{cases}
\]
where $r$ and $p$ are the reward sensitivity and punishment sensitivity parameters, respectively, and $\mathbf{s}(t)$ is a $3 \times 1$ vector denoting the signal amplitude. 
For each generic element $s_{i}(t)$, its value is given by
\[
s_{i}(t) = 
\begin{cases} 
\frac{m_{k,i}(t) a_{i}(t)}{\textbf{m}_k(t)^\top \mathbf{a}(t)} & \text{if } \text{`correct'}, \\
\frac{(1 - m_{k,i}(t))a_{i}(t)}{(\mathbf{1}^\top -\textbf{m}_k(t)^\top) \mathbf{a}(t)} & \text{if } \text{`incorrect'}.
\end{cases}
\]

\section*{Supplementary note 3: Robustness checks}
In total, we examined 19 variants in both the Iowa Gambling Task and Cambridge Gambling Task, and 15 variants in the Wisconsin Card Sorting Task. The variants included: (i) adjusting the LLMs' temperature from the default values to 0 and 0.5; (ii) applying arithmetic transformations to the scores of each round, such as multiplying by a constant factor or adding a constant value (specifically, 0.5x, 2x, +100, +200 for Iowa Gambling Task; 5x, 10x, +100, +200 for Cambridge Gambling Task; and not applicable to Wisconsin Card Sorting Task, as it does not involve scores); (iii) restructuring the prompts to reflect different decision-making contexts, including both economic and medical scenarios; and (iv) introducing role-play prompts to assess the impact of demographic information (e.g., different age ranges, ethnicity, gender, and other categories such as elder, kid, American, Asian, Black, Hispanic, and White) and risk preferences (e.g., risk-taking and risk-averse behavior).

Each variant was repeated 10 times using GPT-4o. Overall, we observed that the decision-making patterns of the LLMs remained qualitatively the same across these variants, confirming the robustness of our findings and highlighting the fundamental differences in decision-making patterns between LLMs and humans. The following summarizes the key observations in these variants.

\subsection*{Iowa Gambling Task}
\begin{figure}
       \centering
       \includegraphics[width=\linewidth]{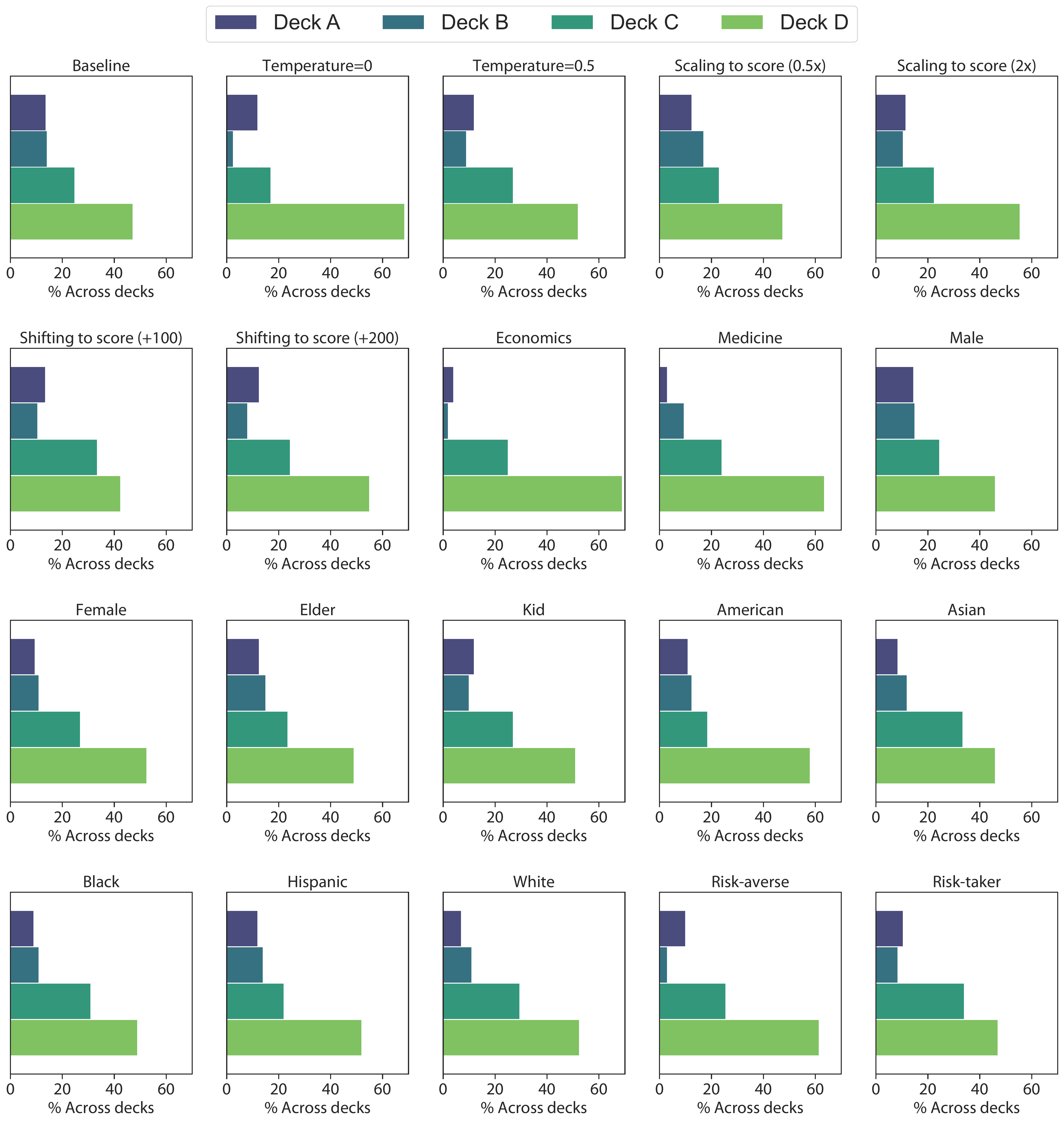}
       \vspace{1em}
       \caption{
            \textbf{Robustness checks on Iowa Gambling Task.} Choice Distribution in the Last 20 Rounds for different prompt variations, using GPT-4o over 10 sessions. The LLM predominantly selected the two advantageous decks across all experimental variants, exhibiting a stronger preference for deck D, which imposed penalties less frequently.
       }
       \label{fig:igt_gpt_4o_robust}
\end{figure}

As shown in \href{SI.com}{\textit{SI Appendix},  Fig.~\ref{fig:igt_gpt_4o_robust}}, GPT-4o consistently reproduced the card choice pattern observed in the baseline condition reported in the main text, favoring deck D over deck C across all 19 prompt variants. This pattern demonstrates that the model's sensitivity to penalty frequency remained robust across varied prompt designs. These results highlight GPT-4o’s stable decision-making strategy under uncertain conditions.

\subsection*{Cambridge Gambling Task} 
\begin{figure}
       \centering
       \includegraphics[width=\linewidth]{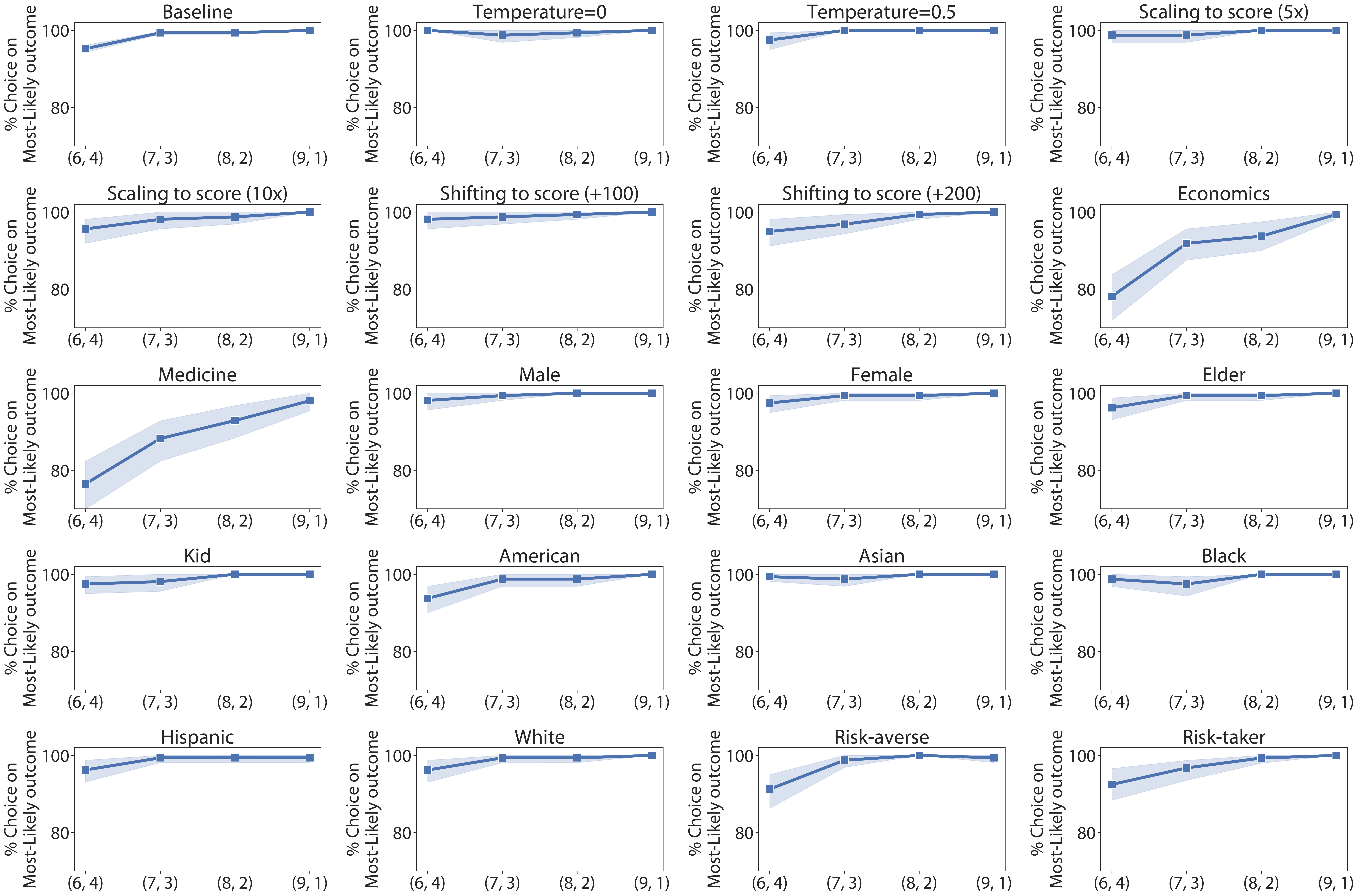}
       \vspace{1em}
       \caption{
            \textbf{Robustness checks for the Cambridge Gambling Task.} Decision-making quality for different prompt  variations, using GPT-4o over 10 sessions. The LLM consistently opted for the majority box type irrespective of the degree of asymmetry in box distributions, thereby reliably predicting the most probable outcome across the various variants.
       }
       \label{fig:cgt_gpt_4o_robust_qm}
\end{figure}

As shown in \href{SI.com}{\textit{SI Appendix}, Fig.~\ref{fig:cgt_gpt_4o_robust_qm}}, the GPT-4o consistently selected the majority box type across all prompt variants, accurately identifying the most probable outcome regardless of asymmetry in box distributions. This pattern held across all variants except for economics and medicine, indicating a broadly robust capacity for probabilistic reasoning.

\begin{figure}
       \centering
       \includegraphics[width=\linewidth]{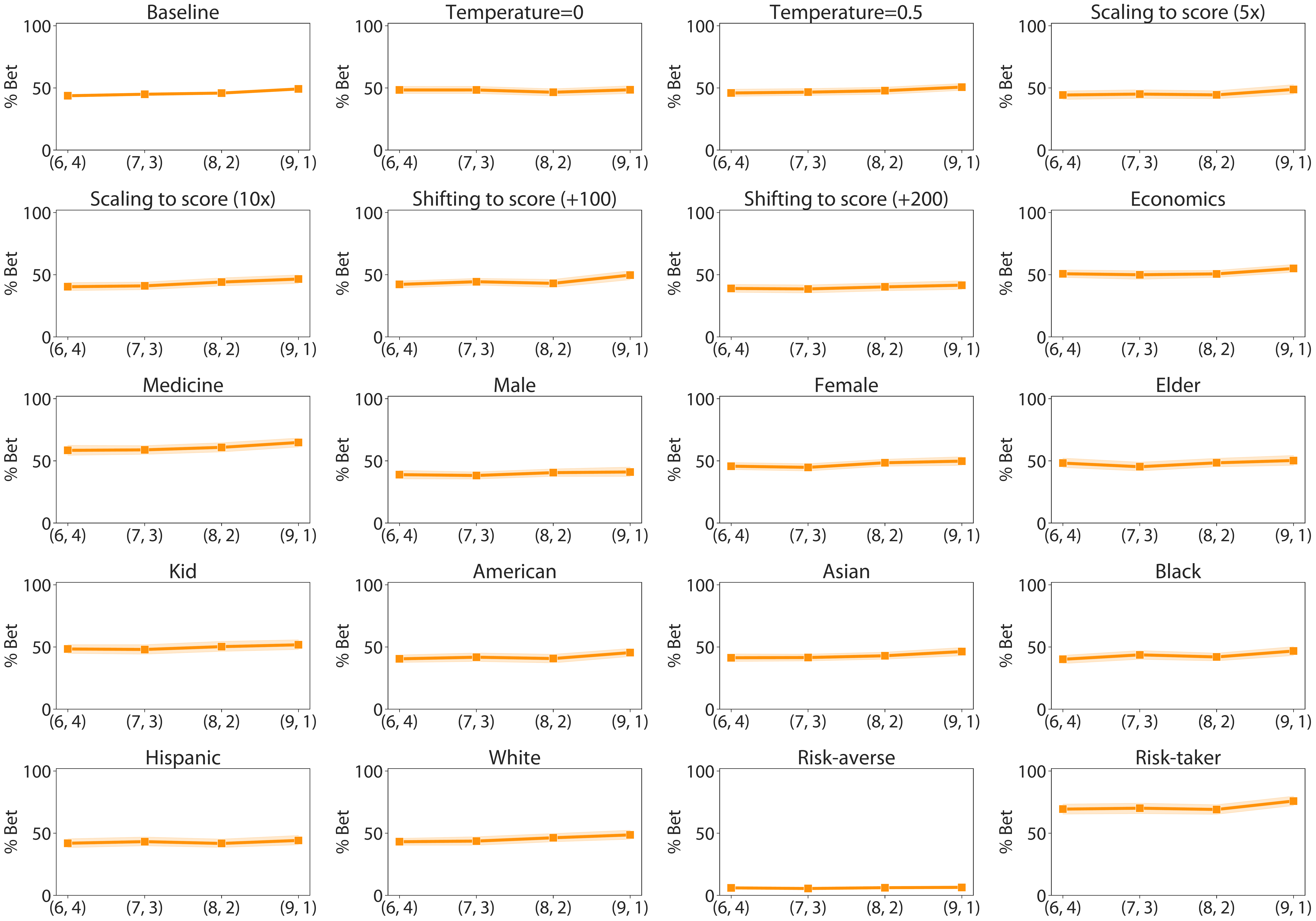}
       \vspace{1em}
       \caption{
            \textbf{Robustness checks for the Cambridge Gambling Task. } Risk adjustment for different prompt variations, using GPT-4o over 10 sessions. The LLM maintained a consistent betting pattern across different levels of asymmetry, indicating a lack of effective risk adjustment. 
       }
       \label{fig:cgt_gpt_4o_robust_ra}
\end{figure}

In contrast, \href{SI.com}{\textit{SI Appendix}, Fig.~\ref{fig:cgt_gpt_4o_robust_ra}} reveals that the GPT-4o exhibited minimal adjustment in its betting behavior across varying levels of risk. Despite increasing asymmetry (e.g., from 6:4 to 9:1 ratios), bet proportions remained largely stable across conditions. Such a lack of sensitivity to shifting odds was consistent across all experimental conditions, which suggests a rigid decision strategy that does not adapt to changing risk levels.

\subsection*{Wisconsin Card Sorting Task} 
\begin{figure}
       \centering
       \includegraphics[width=\linewidth]{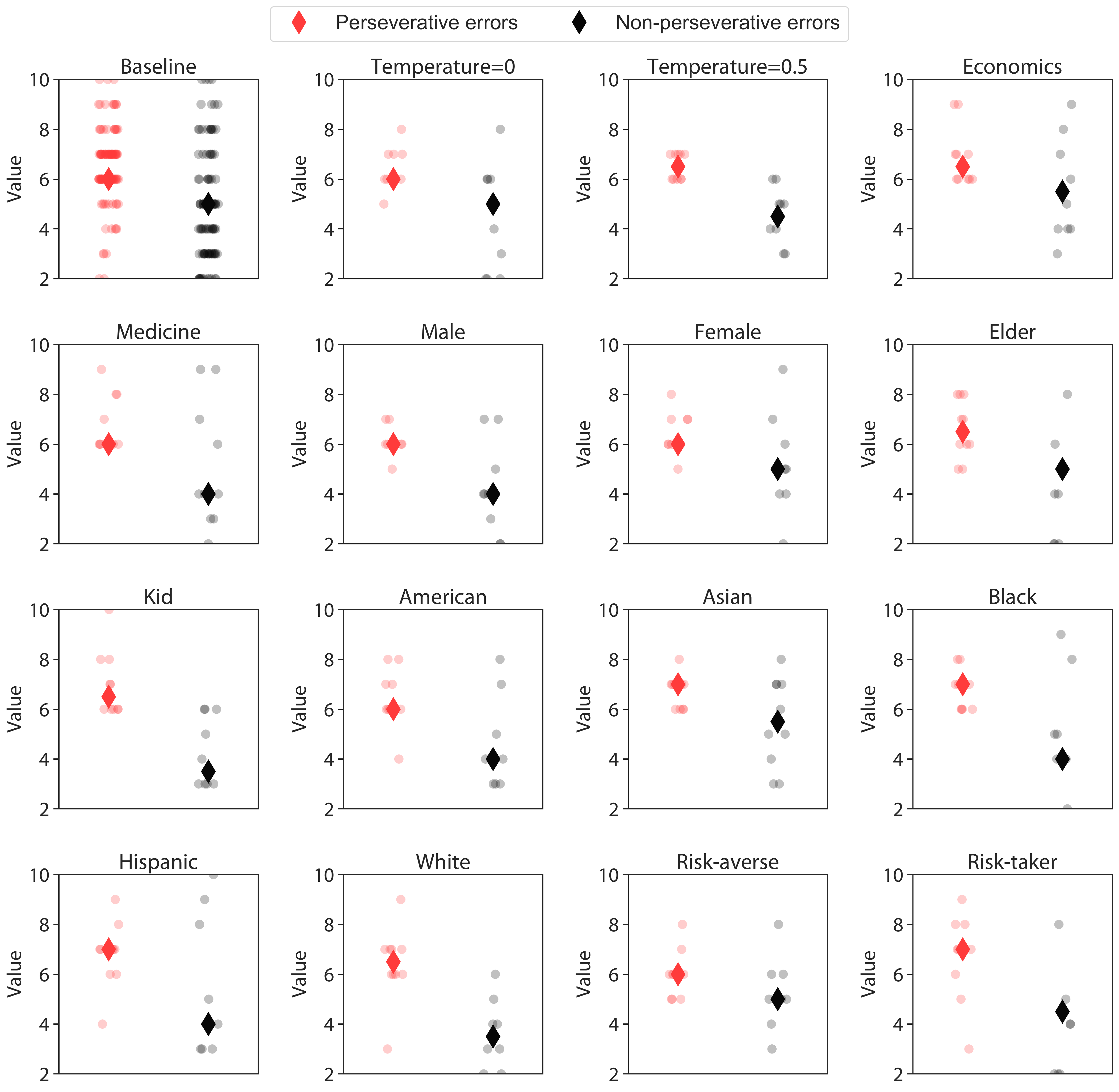}
       \vspace{1em}
       \caption{
            \textbf{Robustness checks for the Wisconsin Card Sorting Task.} Two WCST metrics: Perseverative errors and Non-perseverative errors for different prompt variations, using GPT-4o over 10 sessions. Median values are represented by diamond markers. The LLM demonstrated fewer Non-perseverative errors than Perseverative errors across the variants, thereby evidencing enhanced task proficiency.
       }
       \label{fig:wcst_gpt_4o_robust}
\end{figure}

As shown in \href{SI.com}{\textit{SI Appendix}, Fig.~\ref{fig:wcst_gpt_4o_robust}}, the LLM consistently produced fewer Non-perseverative errors than Perseverative errors across all the variants. The relative reduction in Non-perseverative errors, typically regarded as random or inattentive mistakes, suggests that the LLM maintained strong task focus and demonstrated enhanced rule-learning efficiency and overall performance.

\clearpage

\begin{table*}[!t]
  \centering
  \small
  \caption{Statistical comparisons and parameter estimates for the Iowa Gambling Task}
  \label{tab:igt_results}

  \begin{subtable}[t]{.6\linewidth}
    \centering
    \caption{Comparisons of overall performance between LLMs and humans}
    \setlength{\tabcolsep}{6pt}
    \begin{tabular}{@{}l r r r@{}}
      \toprule
      Model & \multicolumn{1}{c}{$W$} & $p$‑value & Cohen's $d$ \\ \midrule
      GPT      & 2\,556   & $<.001^{***}$ & $-1.135$ \\
      Claude   &   773.5  & $<.001^{***}$ & $-2.456$ \\
      Gemini   & 1\,209   & $<.001^{***}$ & $-1.973$ \\
      GPTo4m   &   815.5  & $<.001^{***}$ & $-2.438$ \\
      DeepSeek &   773    & $<.001^{***}$ & $-2.497$ \\ \bottomrule
    \end{tabular}

    \vspace{0.35em}
    \parbox{\linewidth}{\footnotesize\emph{Note.} Cohen's $d$ represents the standardized difference in net scores between human participants and each LLM, with LLMs serving as the reference group. Negative values indicate that LLMs outperformed humans on the Iowa Gambling Task.}
  \end{subtable}

  \par\bigskip

  \begin{subtable}[t]{\linewidth}
    \centering
    \caption{Within‑group comparisons of deck preferences (Deck C vs. Deck D)}
    \setlength{\tabcolsep}{4pt}
    \begin{tabular}{@{}l r r r r r r r@{}}
      \toprule
      Model  & \# of Deck C & \# of Deck D & \% of Deck C\textsuperscript{*} & \% of Deck D\textsuperscript{*} &
      \multicolumn{1}{c}{$Z$} & $p$-value & Cohen's $h$ \\ \midrule
      Humans     & 388  & 353  & 32.333 & 29.417 &  0.5   & 0.625        &  0.063 \\
      GPT        & 322  & 531  & 26.833 & 44.250 & $-2.8$ & 0.005$^{**}$ & $-0.366$ \\
      Claude     & 111  & 1055 & 9.250 & 87.917 & $-12.2$& $<.001^{***}$& $-1.813$ \\
      Gemini     & 479  & 679  & 39.917 & 56.583 & $-2.6$ & 0.010$^{**}$ & $-0.335$ \\
      GPTo4m     & 963  & 184  & 80.250 & 15.333 & 10.1   & $<.001^{***}$&  1.416 \\
      DeepSeek   & 695  & 472  & 57.917 & 39.333 &  2.9   & 0.004$^{**}$ &  0.374 \\ \bottomrule
    \end{tabular}

    \vspace{0.35em}
    \parbox{\linewidth}{\footnotesize\emph{Note.} Cohen's $h$ reflects the standardized difference in proportions between deck C and deck D selections, computed as a two‑proportion $Z$ test with deck D as the reference ($C-D$). Positive values indicate a greater preference for deck C, while negative values indicate a greater preference for deck D. *The percentage of deck C and D is calculated from the total number of deck A, B, C and D (e.g., $\% \text{ of Deck C} = \frac{\#\text{ of Deck C}}{\#\text{ of Deck A} + \#\text{ of Deck B} + \#\text{ of Deck C} + \#\text{ of Deck D}}$).}
  \end{subtable}

  \vspace{1.3em} 

  \begin{subtable}[t]{.47\linewidth}
    \centering
    \caption{Prospect Valence Learning model with a decay reinforcement learning rule: Pairwise parameter comparisons with human participants}
    \setlength{\tabcolsep}{4.5pt}
    \begin{tabular}{@{}l l r r r@{}}
      \toprule
      Param. & Model & $W$ & $p$‑value & Cohen's $d$ \\ \midrule
      \multirow{5}{*}{$A$}
        & GPT      & 1\,680.0 & $<.001^{***}$ & $-1.696$ \\
        & Claude   &   360.0  & $<.001^{***}$ & $-2.642$ \\
        & Gemini   &    38.0  & $<.001^{***}$ & $-3.113$ \\
        & GPTo4m   &   963.0  & $<.001^{***}$ & $-2.226$ \\
        & DeepSeek &     0.0  & $<.001^{***}$ & $-3.045$ \\ \midrule
      \multirow{5}{*}{$c$}
        & GPT      & 2\,893.0 & $<.001^{***}$ & $-1.283$ \\
        & Claude   & 2\,937.0 & $<.001^{***}$ & $-1.284$ \\
        & Gemini   & 5\,545.0 & 0.002$^{**}$  &  0.004 \\
        & GPTo4m   & 1\,141.0 & $<.001^{***}$ & $-2.288$ \\
        & DeepSeek & 4\,107.0 & $<.001^{***}$ & $-0.511$ \\ \midrule
      \multirow{5}{*}{$\alpha$}
        & GPT      &     0.0  & $<.001^{***}$ & $-7.331$ \\
        & Claude   &     0.0  & $<.001^{***}$ & $-7.167$ \\
        & Gemini   &     0.0  & $<.001^{***}$ & $-6.603$ \\
        & GPTo4m   &     0.0  & $<.001^{***}$ & $-5.849$ \\
        & DeepSeek &     0.0  & $<.001^{***}$ & $-6.058$ \\ \midrule
      \multirow{5}{*}{$\lambda$}
        & GPT      & 6\,720.0 & 0.373         &  0.474 \\
        & Claude   & 4\,674.0 & $<.001^{***}$ & $-0.086$ \\
        & Gemini   & 4\,854.0 & $<.001^{***}$ & $-0.104$ \\
        & GPTo4m   & 4\,063.0 & $<.001^{***}$ & $-0.483$ \\
        & DeepSeek & 3\,480.0 & $<.001^{***}$ & $-0.952$ \\ \bottomrule
    \end{tabular}

    \vspace{0.35em}
    \parbox{\linewidth}{\footnotesize\emph{Note.} Cohen's $d$ represents the standardized difference in parameter estimates between human participants and each LLM, with LLMs serving as the reference group. Negative values indicate that LLMs had higher parameter values than humans.}
  \end{subtable}\hfill
  \begin{subtable}[t]{.47\linewidth}
    \centering
    \caption{Prospect Valence Learning model with a decay reinforcement learning rule: Descriptive statistics for parameters}
    \setlength{\tabcolsep}{4.5pt}
    \begin{tabular}{@{}l l r r r@{}}
      \toprule
      Param. & Model & Mean & Median & SD \\ \midrule
      \multirow[t]{6}{*}{$A$} & Humans & 0.7059 & 0.7028 & 0.1263 \\
         & GPT      & 0.8575 & 0.8579 & 0.0015\\
         & Claude   & 0.942  & 0.9421 & 0.0008\\
         & Gemini   & 0.9869 & 0.9942 & 0.0179\\
         & GPTo4m   & 0.9048 & 0.9049 & 0.0014\\
         & DeepSeek & 0.978  & 0.9782 & 0.0012\\
      \midrule
      \multirow[t]{6}{*}{$c$} & Humans & 0.414  & 0.3484 & 0.2977 \\
         & GPT      & 0.6871 & 0.6893 & 0.0455\\
         & Claude   & 0.6861 & 0.686  & 0.0343\\
         & Gemini   & 0.4132 & 0.4125 & 0.0255\\
         & GPTo4m   & 0.8991 & 0.9022 & 0.0355\\
         & DeepSeek & 0.5216 & 0.5214 & 0.0038\\
      \midrule
      \multirow[t]{6}{*}{$\alpha$} & Humans & 0.3185 & 0.3149 & 0.1219 \\
         & GPT      & 0.9522 & 0.9543 & 0.0089\\
         & Claude   & 0.9366 & 0.937  & 0.0036\\
         & Gemini   & 0.9479 & 0.9448 & 0.0575\\
         & GPTo4m   & 0.8227 & 0.8227 & 0.0013\\
         & DeepSeek & 0.8407 & 0.8408 & 0.0007\\
      \midrule
      \multirow[t]{6}{*}{$\lambda$} & Humans & 1.1005 & 0.5903 & 1.1493 \\
         & GPT      & 0.7151 & 0.7161 & 0.0092\\
         & Claude   & 1.1705 & 1.1708 & 0.0035\\
         & Gemini   & 1.1865 & 1.212  & 0.2196\\
         & GPTo4m   & 1.4933 & 1.4932 & 0.0037\\
         & DeepSeek & 1.874  & 1.8741 & 0.0015\\
        \midrule
    \end{tabular}
  \end{subtable}
\end{table*}

\begin{table*}[!t]
  \centering
  \small
  \caption{Statistical comparisons and parameter estimates for the Cambridge Gambling Task}
  \label{tab:cgt_results}

  \begin{subtable}[t]{.97\linewidth}
    \centering
    \caption{Comparisons of overall performance (Total Scores) between LLMs and humans}
    \setlength{\tabcolsep}{6pt}
    \begin{tabular}{@{}l r r r@{}}
      \toprule
      Model & \multicolumn{1}{c}{$W$} & $p$‑value & Cohen's $d$ \\ \midrule
      GPT      & 6\,354   & 0.116            & $-0.100$ \\
      Claude   & 4\,175   & $<.001^{***}$    & $-0.691$ \\
      Gemini   & 9\,282.5 & $<.001^{***}$    &  0.484 \\
      GPTo4m   & 4\,774   & $<.001^{***}$    & $-1.155$ \\
      DeepSeek & 5\,029   & $<.001^{***}$    & $-1.043$ \\ \bottomrule
    \end{tabular}

    \vspace{0.35em}
    \parbox{\linewidth}{\footnotesize\emph{Note.} Cohen's $d$ represents the standardized difference in total scores between human participants and each LLM, with LLMs serving as the reference group. Negative values indicate that LLMs outperformed humans on the Cambridge Gambling Task.}
  \end{subtable}

  \vspace{1.3em}

  \begin{subtable}[t]{.47\linewidth}
    \centering
    \caption{Cumulative model: Pairwise parameter comparisons with humans}
    \setlength{\tabcolsep}{4.5pt}
    \begin{tabular}{@{}l l r r r@{}}
      \toprule
      Param. & Model & $W$ & $p$‑value & Cohen's $d$ \\ \midrule
      \multirow{5}{*}{$c$}
        & GPT      & 1\,071.0 & $<.001^{***}$ & $-1.786$ \\
        & Claude   & 8\,251.0 & 0.051         &  0.174   \\
        & Gemini   &   121.0  & $<.001^{***}$ & $-4.207$ \\
        & GPTo4m   & 4\,811.0 & $<.001^{***}$ & $-0.108$ \\
        & DeepSeek & 12\,900.0& $<.001^{***}$ &  1.564   \\ \midrule
      \multirow{5}{*}{$\alpha$}
        & GPT      &     0.0  & $<.001^{***}$ & $-2.068$ \\
        & Claude   &   389.0  & $<.001^{***}$ & $-1.710$ \\
        & Gemini   & 1\,864.0 & $<.001^{***}$ & $-1.156$ \\
        & GPTo4m   &     0.0  & $<.001^{***}$ & $-2.078$ \\
        & DeepSeek &   410.0  & $<.001^{***}$ & $-1.771$ \\ \midrule
      \multirow{5}{*}{$\rho$}
        & GPT      & 2\,940.0 & $<.001^{***}$ & $-0.885$ \\
        & Claude   & 12\,992.0& $<.001^{***}$ &  1.826   \\
        & Gemini   &     0.0  & $<.001^{***}$ & $-2.610$ \\
        & GPTo4m   & 14\,400.0& $<.001^{***}$ &  4.346   \\
        & DeepSeek & 14\,400.0& $<.001^{***}$ &  5.466   \\ \midrule
      \multirow{5}{*}{$\gamma$}
        & GPT      & 12\,697.0& $<.001^{***}$ &  1.299   \\
        & Claude   & 12\,236.0& $<.001^{***}$ &  1.174   \\
        & Gemini   & 11\,560.0& $<.001^{***}$ &  1.088   \\
        & GPTo4m   &   564.0  & $<.001^{***}$ & $-2.201$ \\
        & DeepSeek &   218.0  & $<.001^{***}$ & $-1.068$ \\ \bottomrule
    \end{tabular}

    \vspace{0.35em}
    \parbox{\linewidth}{\footnotesize\emph{Note.} Cohen's $d$ represents the standardized difference in parameter estimates between human participants and each LLM, with LLMs serving as the reference group. Negative values indicate that LLMs had higher parameter values than humans.}
  \end{subtable}\hfill
  \begin{subtable}[t]{.47\linewidth}
    \centering
    \caption{Cumulative model: Descriptive statistics for parameters}
    \setlength{\tabcolsep}{4.5pt}
    \begin{tabular}{@{}l l r r r@{}}
      \toprule
      Param. & Model & Mean & Median & SD \\ \midrule
      \multirow{6}{*}{$c$}
        & Humans   & 0.4955 & 0.4961 & 0.0301 \\
        & GPT      & 0.5338 & 0.5331 & 0.0035 \\
        & Claude   & 0.4908 & 0.4918 & 0.0225 \\
        & Gemini   & 0.5866 & 0.5842 & 0.0057 \\
        & GPTo4m   & 0.4978 & 0.4978 & 0.0002 \\
        & DeepSeek & 0.4617 & 0.4627 & 0.0051 \\ \midrule
      \multirow{6}{*}{$\alpha$}
        & Humans   & 2.9341 & 3.1058 & 1.4003 \\
        & GPT      & 4.9822 & 4.9822 & 0.0002 \\
        & Claude   & 4.7595 & 4.8839 & 0.5637 \\
        & Gemini   & 4.3296 & 4.8154 & 0.9767 \\
        & GPTo4m   & 4.9915 & 4.9915 & 0.0001 \\
        & DeepSeek & 4.8124 & 4.9277 & 0.5382 \\ \midrule
      \multirow{6}{*}{$\rho$}
        & Humans   &  63.8021 &   6.4272 & 88.4852 \\
        & GPT      & 418.6155 & 512.7302 & 306.6256 \\
        & Claude   &   0.7864 &   0.3560 &  1.0657 \\
        & Gemini   & 691.8412 & 692.5117 &  5.4954 \\
        & GPTo4m   &   0.0104 &   0.0088 &  0.0056 \\
        & DeepSeek &   0.0018 &   0.0018 &  0.0000 \\ \midrule
      \multirow{6}{*}{$\gamma$}
        & Humans   &  15.8826 & 12.5300 & 11.2113 \\
        & GPT      &   5.5439 &  5.4863 &  1.0043 \\
        & Claude   &   6.5740 &  6.5812 &  0.0470 \\
        & Gemini   &   7.1626 &  7.1715 &  1.6764 \\
        & GPTo4m   &  48.7207 & 44.7819 & 17.8695 \\
        & DeepSeek & 125.3505 & 65.0248 &144.5221 \\ \bottomrule
    \end{tabular}
  \end{subtable}
\end{table*}

\begin{table*}[!t]
  \centering
  \small
  \caption{Statistical comparisons and parameter estimates for the Wisconsin Card Sorting Task}
  \label{tab:wcst_results}

  \begin{subtable}[t]{.47\linewidth}
    \centering
    \caption{Comparisons of overall performance between LLMs and humans}
    \setlength{\tabcolsep}{6pt}
    \begin{tabular}{@{}l r r r@{}}
      \toprule
      Model & \multicolumn{1}{c}{$W$} & $p$‑value & Cohen's $d$ \\ \midrule
      GPT      & 4\,910.5 & $<.001^{***}$ & $-0.399$ \\
      Claude   & 3\,579.0 & $<.001^{***}$ & $-0.925$ \\
      Gemini   & 3\,109.5 & $<.001^{***}$ & $-0.934$ \\
      GPTo4m   & 2\,854.0 & $<.001^{***}$ & $-1.218$ \\
      DeepSeek & 7\,391.5 & 0.722         &  0.332  \\ \bottomrule
    \end{tabular}

    \vspace{0.35em}
    \parbox{\linewidth}{\footnotesize\emph{Note.} Cohen's $d$ represents the standardized difference in overall performance between human participants and each LLM, with LLMs serving as the reference group. Negative values indicate that LLMs outperformed humans on the Wisconsin Card Sorting Task.}
  \end{subtable}\hfill
  \begin{subtable}[t]{.47\linewidth}
    \centering
    \caption{Within‑group comparisons of error patterns (perseverative vs.\ non‑perseverative)}
    \setlength{\tabcolsep}{6pt}
    \begin{tabular}{@{}l r r r@{}}
      \toprule
      Model & \multicolumn{1}{c}{$W$} & $p$‑value & Cohen's $d$ \\ \midrule
      Humans   & 4\,821.0  & $<.001^{***}$ & $-0.868$ \\
      GPT      & 9\,266.5  & $<.001^{***}$ & $-0.030$ \\
      Claude   &11\,079.5  & $<.001^{***}$ &  0.611 \\
      Gemini   &12\,124.5  & $<.001^{***}$ &  0.480 \\
      GPTo4m   &12\,636.5  & $<.001^{***}$ &  1.195 \\
      DeepSeek & 5\,040.5  & $<.001^{***}$ & $-1.013$ \\ \bottomrule
    \end{tabular}

    \vspace{0.35em}
    \parbox{\linewidth}{\footnotesize\emph{Note.} Cohen's $d$ reflects the standardized difference between perseverative errors and non‑perseverative errors. Negative values indicate more non‑perseverative errors relative to perseverative errors.}
  \end{subtable}

  \vspace{1.3em}

  \begin{subtable}[t]{.47\linewidth}
    \centering
    \caption{Sequential learning model: Pairwise parameter comparisons with humans}
    \setlength{\tabcolsep}{4.5pt}
    \begin{tabular}{@{}l l r r r@{}}
      \toprule
      Param. & Model & $W$ & $p$‑value & Cohen's $d$ \\ \midrule
      \multirow{5}{*}{$r$}
        & GPT      &     0.0  & $<.001^{***}$ & $-18.875$ \\
        & Claude   &     0.0  & $<.001^{***}$ & $-17.957$ \\
        & Gemini   &     0.0  & $<.001^{***}$ & $-16.902$ \\
        & GPTo4m   &     0.0  & $<.001^{***}$ & $-18.696$ \\
        & DeepSeek & 14\,400.0& $<.001^{***}$ &  16.079  \\ \midrule
      \multirow{5}{*}{$p$}
        & GPT      & 4\,333.0 & $<.001^{***}$ & $-0.263$ \\
        & Claude   & 2\,859.0 & $<.001^{***}$ & $-0.294$ \\
        & Gemini   &     0.0  & $<.001^{***}$ & $-1.238$ \\
        & GPTo4m   &     0.0  & $<.001^{***}$ & $-1.209$ \\
        & DeepSeek &     0.0  & $<.001^{***}$ & $-1.241$ \\ \midrule
      \multirow{5}{*}{$d$}
        & GPT      & 5\,132.0 & $<.001^{***}$ & $-0.470$ \\
        & Claude   & 2\,973.0 & $<.001^{***}$ & $-1.171$ \\
        & Gemini   & 2\,963.0 & $<.001^{***}$ & $-1.080$ \\
        & GPTo4m   & 2\,219.0 & $<.001^{***}$ & $-1.433$ \\
        & DeepSeek & 7\,602.0 & 0.455         & $-0.253$ \\ \bottomrule
    \end{tabular}

    \vspace{0.35em}
    \parbox{\linewidth}{\footnotesize\emph{Note.} Cohen's $d$ represents the standardized difference in parameter estimates between human participants and each LLM, with LLMs serving as the reference group. Negative values indicate that LLMs had higher parameter values than humans.}
  \end{subtable}\hfill
  \begin{subtable}[t]{.47\linewidth}
    \centering
    \caption{Sequential learning model: Descriptive statistics for parameters}
    \setlength{\tabcolsep}{4.5pt}
    \begin{tabular}{@{}l l r r r@{}}
      \toprule
      Param. & Model & Mean & Median & SD \\ \midrule
      \multirow{6}{*}{$r$}
        & Humans   & 0.9974 & 0.9975 & 0.0002 \\
        & GPT      & 0.9997 & 0.9997 & 0.0000 \\
        & Claude   & 0.9996 & 0.9996 & 0.0000 \\
        & Gemini   & 0.9995 & 0.9995 & 0.0000 \\
        & GPTo4m   & 0.9997 & 0.9997 & 0.0000 \\
        & DeepSeek & 0.9729 & 0.9727 & 0.0022 \\ \midrule
      \multirow{6}{*}{$p$}
        & Humans   & 0.9238 & 0.9548 & 0.0867 \\
        & GPT      & 0.9486 & 0.9772 & 0.1010 \\
        & Claude   & 0.9531 & 0.9895 & 0.1108 \\
        & Gemini   & 0.9997 & 0.9997 & 0.0000 \\
        & GPTo4m   & 0.9979 & 0.9980 & 0.0002 \\
        & DeepSeek & 0.9999 & 0.9999 & 0.0000 \\ \midrule
      \multirow{6}{*}{$d$}
        & Humans   & 0.3626 & 0.3563 & 0.1454 \\
        & GPT      & 0.4308 & 0.4519 & 0.1449 \\
        & Claude   & 0.5496 & 0.5436 & 0.1728 \\
        & Gemini   & 0.5098 & 0.5149 & 0.1266 \\
        & GPTo4m   & 0.5507 & 0.5513 & 0.1153 \\
        & DeepSeek & 0.4392 & 0.2924 & 0.4037 \\ \bottomrule
    \end{tabular}
  \end{subtable}
\end{table*}

\clearpage
\begin{table}\centering
  \caption{Summary of participant demographics by task and session. IGT denotes Iowa Gambling Task, CGT denotes Cambridge Gambling Task, and WCST denotes Wisconsin Card Sorting Task.}

  \begin{tabular}{llllllll}
  \hline
  Date         & Task & Location & Rounds & Participants & Mean age & SD age & \%Women \\ \hline
  14 Jul 2024  & IGT       & Xi'an    & 80     & 40           & 19.6     & 1.6    & 42.5    \\
  20 Jul 2024  & IGT       & Xi'an    & 80     & 40           & 20.7     & 1.5    & 50.0    \\
  20 Jul 2024  & IGT       & Xi'an    & 80     & 40           & 20.1     & 1.5    & 55.0    \\
  
  
  7 Sept 2024  & CGT       & Xi'an    & 64     & 40           & 18.4     & 1.3    & 35.0    \\
  7 Sept 2024  & CGT       & Xi'an    & 64     & 40           & 18.7     & 2.0    & 32.5    \\
  7 Sept 2024  & CGT       & Xi'an    & 64     & 40           & 18.2     & 1.9    & 27.5    \\ 
  
  7 Dec  2024  & WCST      & Xi'an  & 64     & 30           & 18.3     & 0.5    & 36.7    \\
  7 Dec  2024  & WCST      & Xi'an  & 64     & 30           & 18.5     & 0.8    & 23.3    \\
  8 Dec  2024  & WCST      & Xi'an  & 64     & 30           & 18.1     & 0.7    & 46.7    \\
  8 Dec  2024  & WCST      & Xi'an  & 64     & 30           & 18.3     & 0.7    & 23.3    \\  \hline
  \end{tabular}

  \label{table:human_information}
\end{table}

\begin{table}\centering
\caption{The parameters in computational models.}
\begin{tabular}{cccc}
\hline
Model                                                                   & Parameter & Range of Values         & Interpretation                                   \\ \hline
\multirow{4}{*}{PVL-DecayRI model~\cite{ahn2008comparison}}             & $A$       & $0 \leq A \leq 1$       & Learning rate        \\
                                                                        & $c$       & $0 \leq c \leq 5$       & Choice consistency                             \\
                                                                        & $\alpha$  & $0 \leq \alpha \leq 2$  & Outcome sensitivity \\
                                                                        & $\lambda$ & $0 \leq \lambda\leq 10$ & Loss aversion \\ \hline
                                                                        
\multirow{4}{*}{Cumulative model~\cite{romeu2020computational}}         & $c$       & $0 \leq c \leq 1$       & Type bias for RED                   \\
                                                                        & $\alpha$  & $0 \leq \alpha \leq 5$  & Probability distortion              \\
                                                                        & $\rho$    & $0 \leq \rho<+\infty$   & Risk aversion                                 \\
                                                                        & $\gamma$  & $0 \leq \gamma<+\infty$ & Choice consistency                             \\ \hline
                                                                        
\multirow{3}{*}{Sequential learning model~\cite{bishara2010sequential}} & $r$       & $0 \leq r \leq 1$       & Reward sensitivity                               \\
                                                                        & $p$       & $0 \leq p \leq 1$       & Punishment sensitivity                           \\
                                                                        & $d$       & $0 \leq d \leq 5$       & Choice consistency                             \\ \hline
\end{tabular}
\label{table:Computation_model_info}
\end{table}
\clearpage

\begin{figure}
       \centering
       \includegraphics[width=0.6\linewidth]{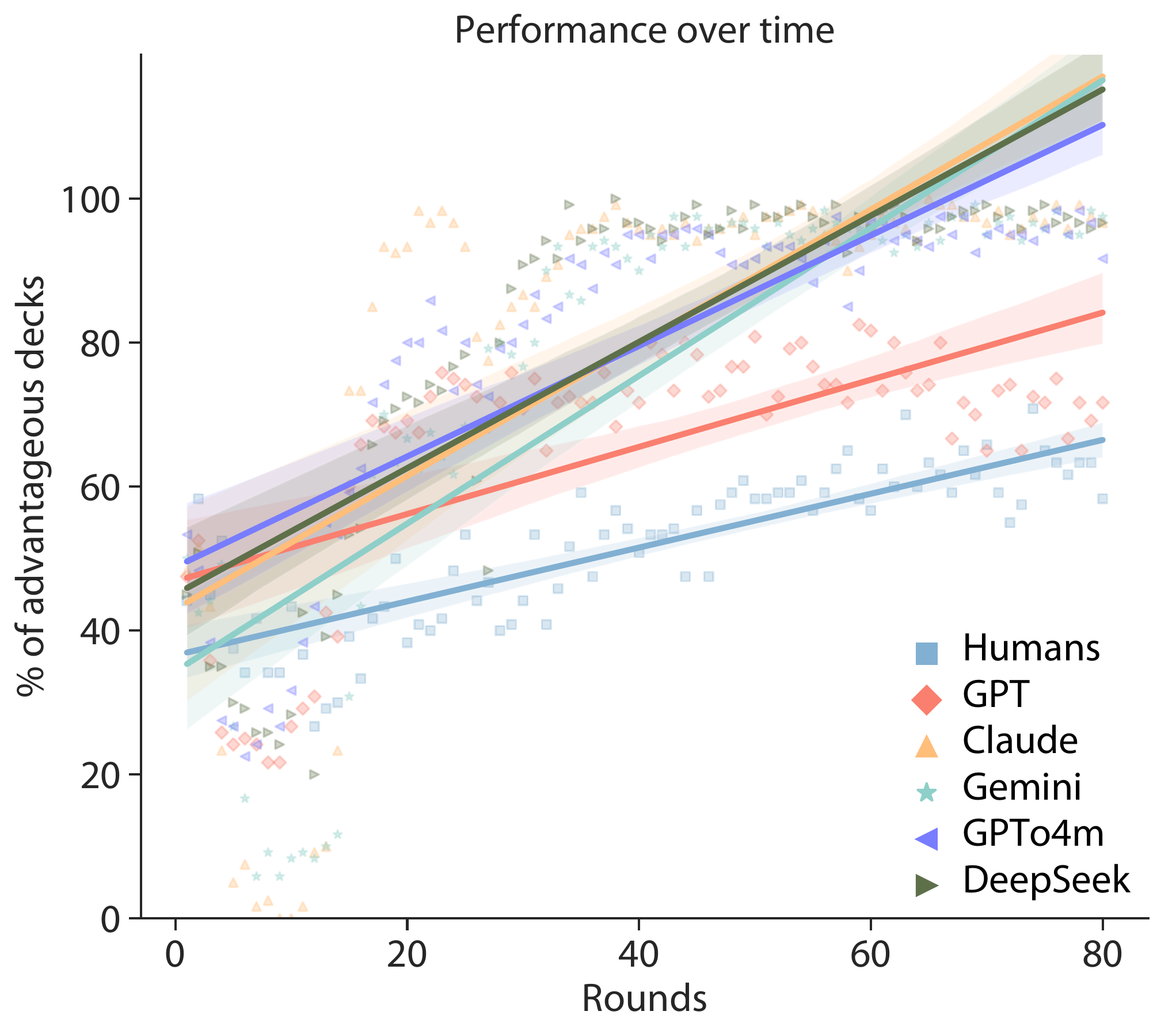}
       \vspace{1em}
       \caption{\textbf{LLMs learned faster than humans in the Iowa Gambling Task, showing steeper increases in advantageous deck selections over time.} The plot shows the proportion of advantageous choices (decks C or D) across 80 rounds. Human participants ($\text{slope}=0.374$, $r=0.83$, $p<.001$), GPT ($\text{slope}=0.466$, $r=0.648$, $<.001$), Claude ($\text{slope}=0.925$, $r=0.69$, $p<.001$), GPTo4m ($\text{slope}=0.768$, $r=0.813$, $p<.001$), Gemini ($\text{slope}=1.027$, $r=0.812$, $p<.001$), and \textcolor{black}{DeepSeek ($\text{slope}=0.877$, $r=0.822$, $p<.001$)} all exhibited a significant increasing trend of choosing advantageous decks, with LLMs demonstrating faster learning speeds compared to humans.}

       \label{fig:IGT_fig1_b}
\end{figure}

\begin{figure}
       \centering
       \includegraphics[width=0.6\linewidth]{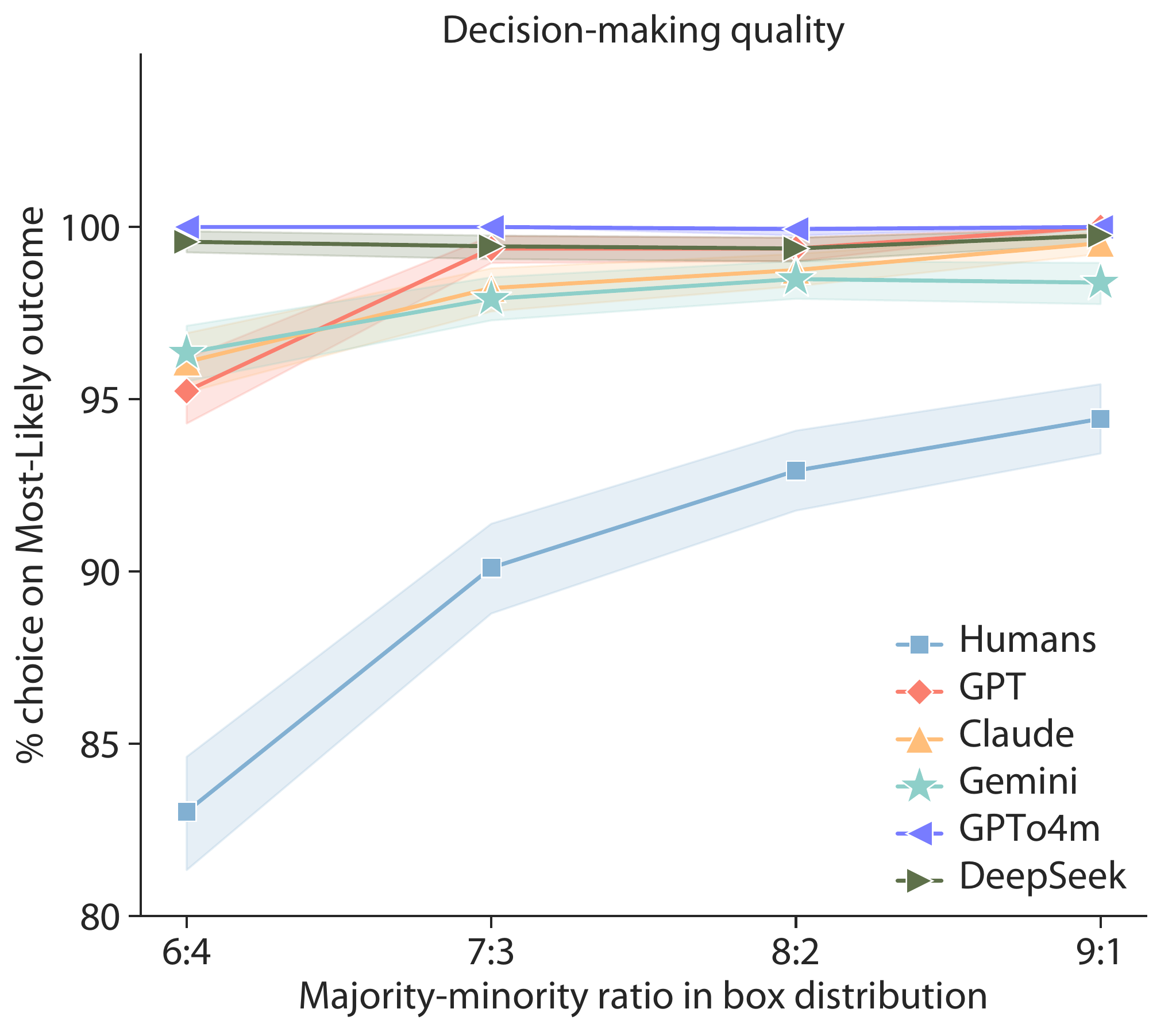}
       \vspace{1em}
       \caption{\textbf{LLMs demonstrated consistently higher decision-making quality than humans across all levels of risk conditions.} This panel shows the decision-making quality in terms of the percentage of selecting the most likely outcome. Human performance improved with increasing asymmetry, but remained below that of all LLMs, which approached ceiling-level accuracy across conditions.}
       \label{fig:CGT_fig1_b}
\end{figure}

\begin{figure}
       \centering
       \includegraphics[width=0.8\linewidth]{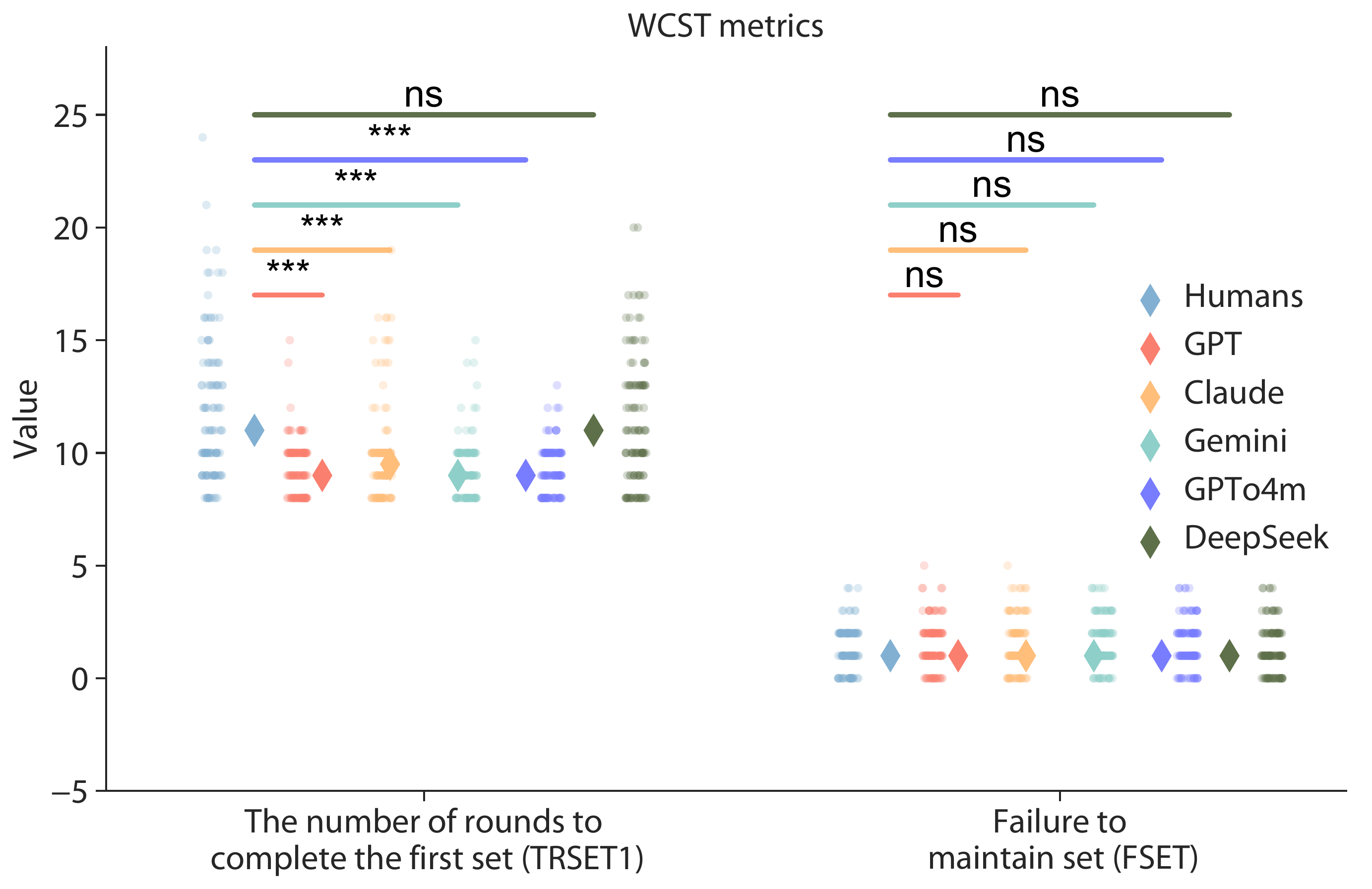}
       \vspace{1em}
       \caption{\textbf{LLMs identified the correct matching rule more quickly than humans but showed similar stability in maintaining the correct strategy.} TRSET1 in the left panel represents the number of rounds to complete the first set. LLMs required significantly fewer rounds compared to human participants (
       GPT: $W=10{,}612$, $p<.001$, Cohen's\ $d=0.976$; 
       Claude: $W=9{,}828.5$, $p<.001$, Cohen's\ $d=0.644$; 
       Gemini: $W=10{,}809$, $p<.001$, Cohen's\ $d=0.982$; \textcolor{black}{
       GPTo4m: $W=10{,}756$, $p<.001$, Cohen's $d=1.018$; nevertheless, 
       DeepSeek ($W=7{,}627$, $p=0.423$, Cohen's\ $d=0.128$) matched human participants.)}. 
       In the right panel, FSET measures how often participants changed their matching strategy after achieving five or more consecutive correct matches without negative feedback (`incorrect'). Human participants and LLMs exhibited comparable FSET values (
       GPT: $W=6{,}553.5$, $p=0.208$, Cohen's\ $d=-0.189$; 
       Claude: $W=6{,}906.5$, $p=0.569$, Cohen's\ $d=-0.142$; 
       Gemini: $W=6{,}443$, $p=0.1414$, Cohen's\ $d=-0.236$;
       GPTo4m: $W=6{,}698$, $p=0.3276$, Cohen's $d=-0.16$; 
       DeepSeek: $W=7{,}528$, $p=0.5241$, Cohen's $d=0.033$).}
       \label{fig:WCST_fig2_TRSET1_FSET}
\end{figure}

\begin{figure}
\centering
\includegraphics[width=0.8\textwidth]{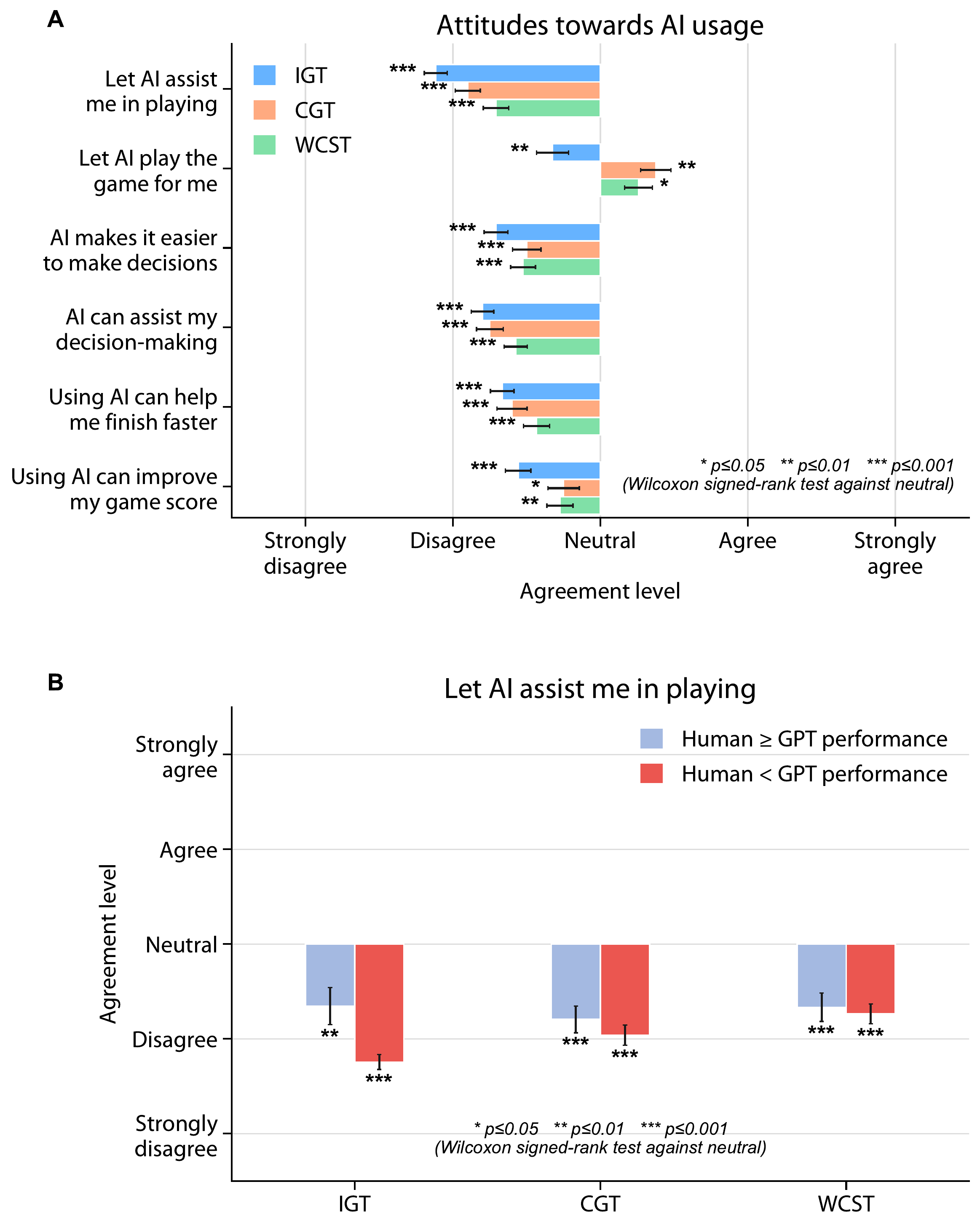}
\vspace{1em}
\caption{\textbf{Participants generally exhibit an overall negative attitude toward AI assistance across all tasks.} \textbf{Panel (A)} shows the participants significantly disagreed with statements reflecting positive perceptions of AI, including believing AI can assist their decision-making (
IGT: $V = 546.5$, $p < .001$; 
CGT: $V = 829.5$, $p < .001$; 
WCST: $V = 590.0$, $p < .001$), believing AI makes it easier to make decisions 
(IGT: $V = 575.0$, $p < .001$; CGT: $V = 1{,}113.0$, $p < .001$; WCST: $V = 780.0$, $p < .001$), 
believing AI can help them finish faster (
IGT: $V = 624.5$, $p < .001$; 
CGT: $V = 1{,}275.0$,$p < .001$; 
WCST: $V = 801.0$, $p < .001$), believing using AI can improve their game scores (
IGT: $V = 774.0$, $p < .001$; 
CGT: $V = 1{,}598.5$, $p = 0.0181$; 
WCST: $V = 932.0$, $p = 0.0034$), and letting AI assist them in playing (
IGT: $V = 335.5$, $p < .001$; 
CGT: $V = 808.5$, $p < .001$; 
WCST: $V = 1{,}028.0$, $p < .001$). 
Conversely, participants displayed relatively neutral or slightly positive attitudes toward letting AI play the game for them (
IGT: $V = 1{,}557.0$, $p = 0.0032$;
CGT: $V = 1{,}563.0$, $p = 0.0012$; 
WCST: $V = 1{,}470.0$, $p = 0.0126$). 
\textbf{Panel (B)} shows that participants expressed reluctance to accept AI assistance, regardless of whether GPT outperformed them. After each task, participants rated their agreement with the statement ``Let AI assist me in playing.'' Responses were split based on whether the participant's performance was equal to or better than GPT (blue), or worse than GPT (red). We found that participants maintained negative attitudes toward AI assistance even when informed of the LLMs' high performance on the same tasks and despite their own lower scores (
IGT: $V = 115.0$, $p < .001$; 
CGT: $V = 317.0$, $p < .001$; 
WCST: $V = 361.5$, $p < .001$).
Statistical significance from neutrality was assessed using the Wilcoxon signed-rank test.
Bars represent the mean agreement level, with 95\% confidence intervals of the mean shown as error bars.
}
\label{fig:ai_attitude}
\end{figure}

\begin{figure}
\centering
\includegraphics[width=0.8\textwidth]{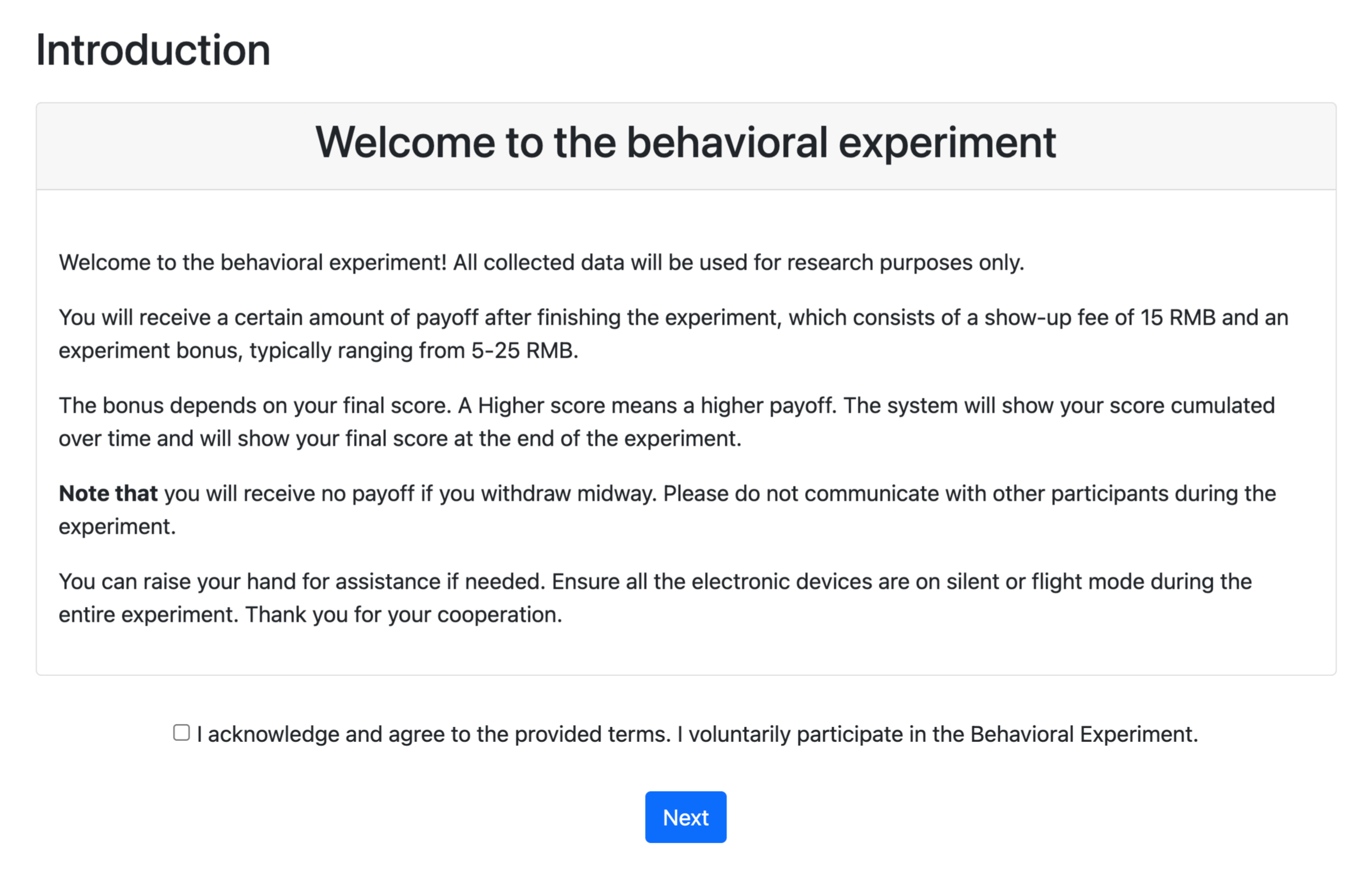}
\includegraphics[width=0.8\textwidth]{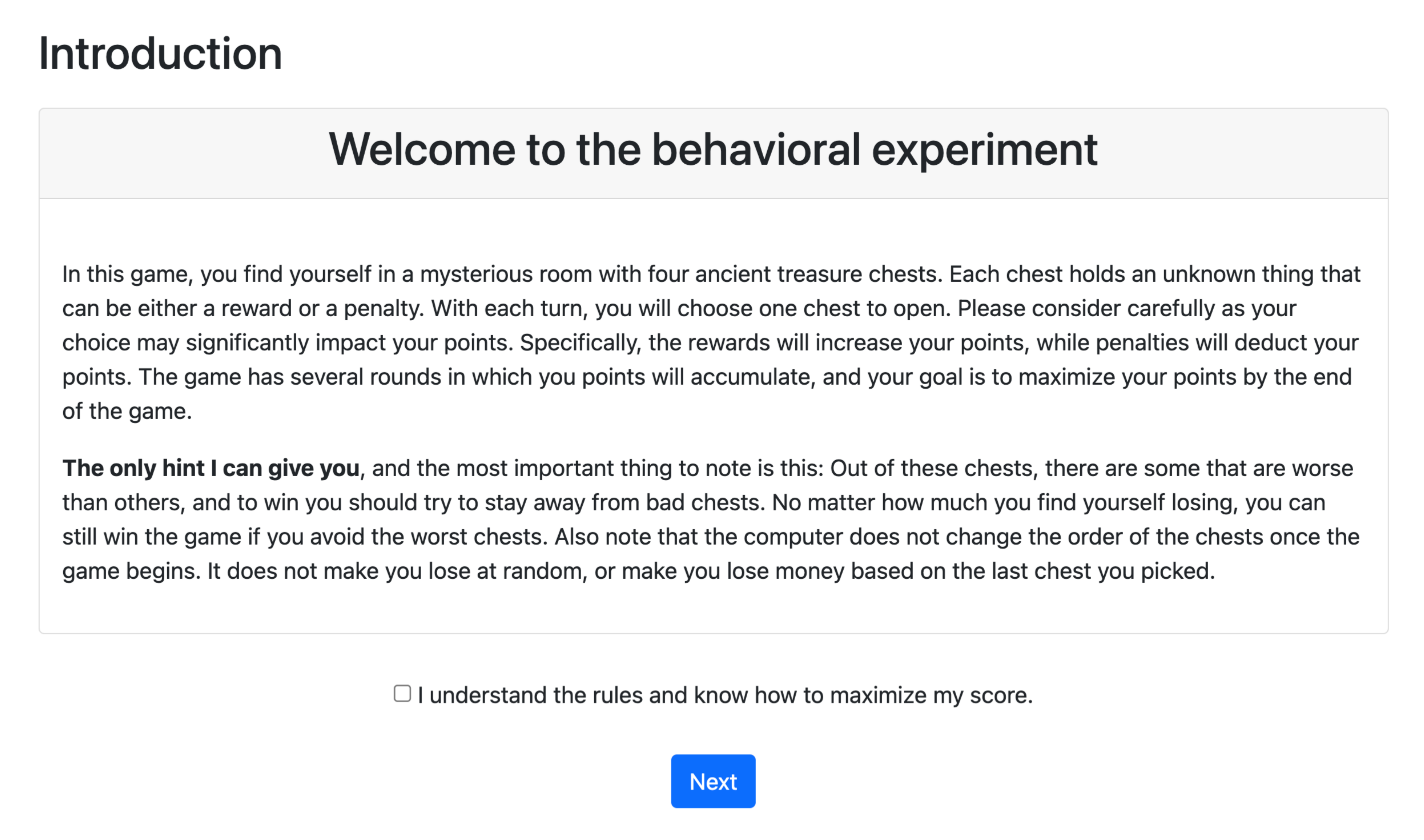}
\vspace{1em}
\caption{Pages of introduction on Iowa Gambling Task. Once participants have understood the task, they can begin by clicking the "Next" button.}
\label{fig:intro_of_igt}
\end{figure}

\begin{figure}
\centering
\includegraphics[width=0.8\textwidth]{SI_figs/pre_intro.pdf}
\includegraphics[width=0.8\textwidth]{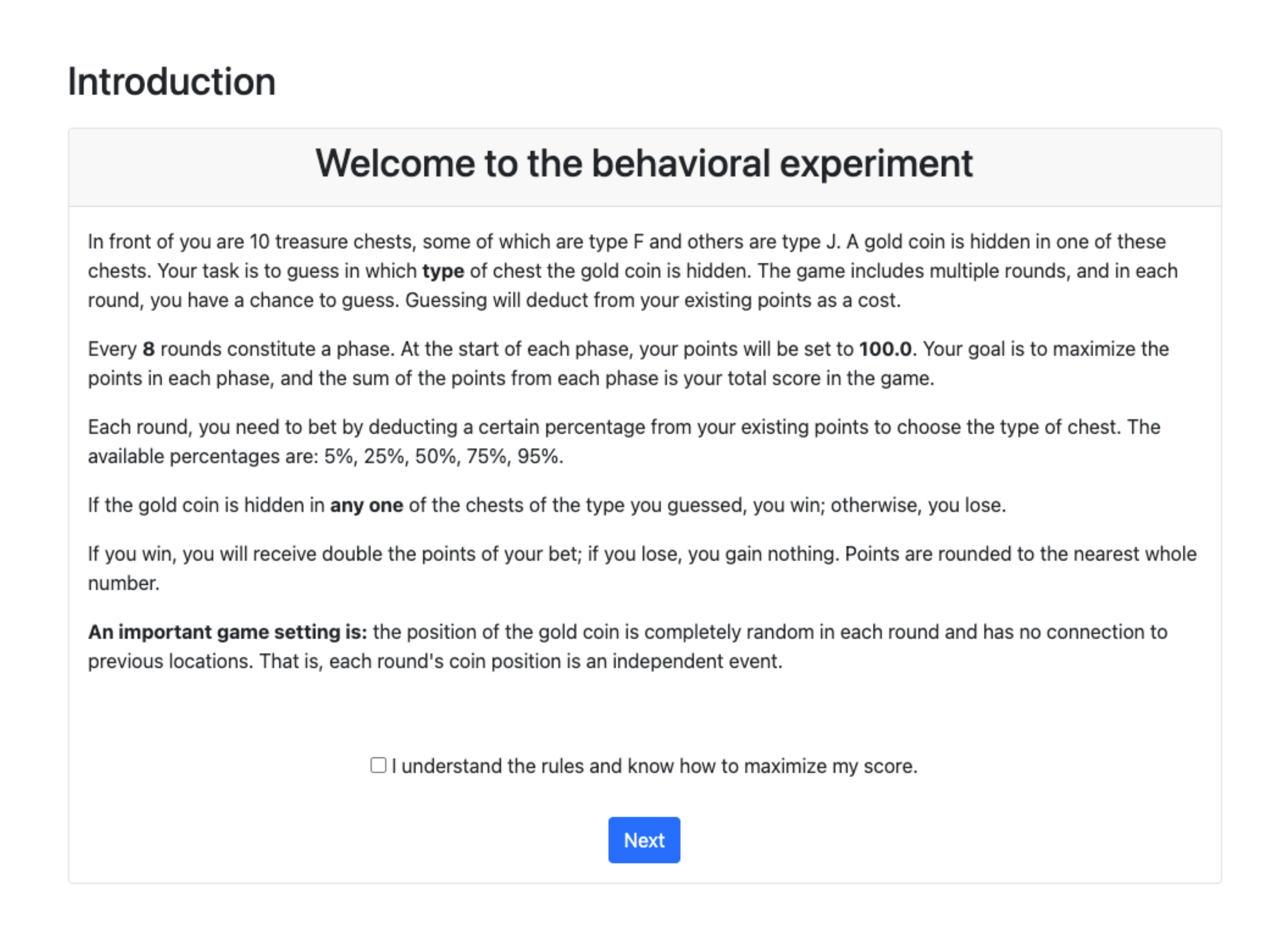}
\vspace{1em}
\caption{Pages of introduction on Cambridge Gambling Task. Once participants have understood the task, they can begin by clicking the "Next" button.}
\label{fig:intro_of_cgt}
\end{figure}

\begin{figure}
\centering
\includegraphics[width=0.8\textwidth]{SI_figs/pre_intro.pdf}
\includegraphics[width=0.8\textwidth]{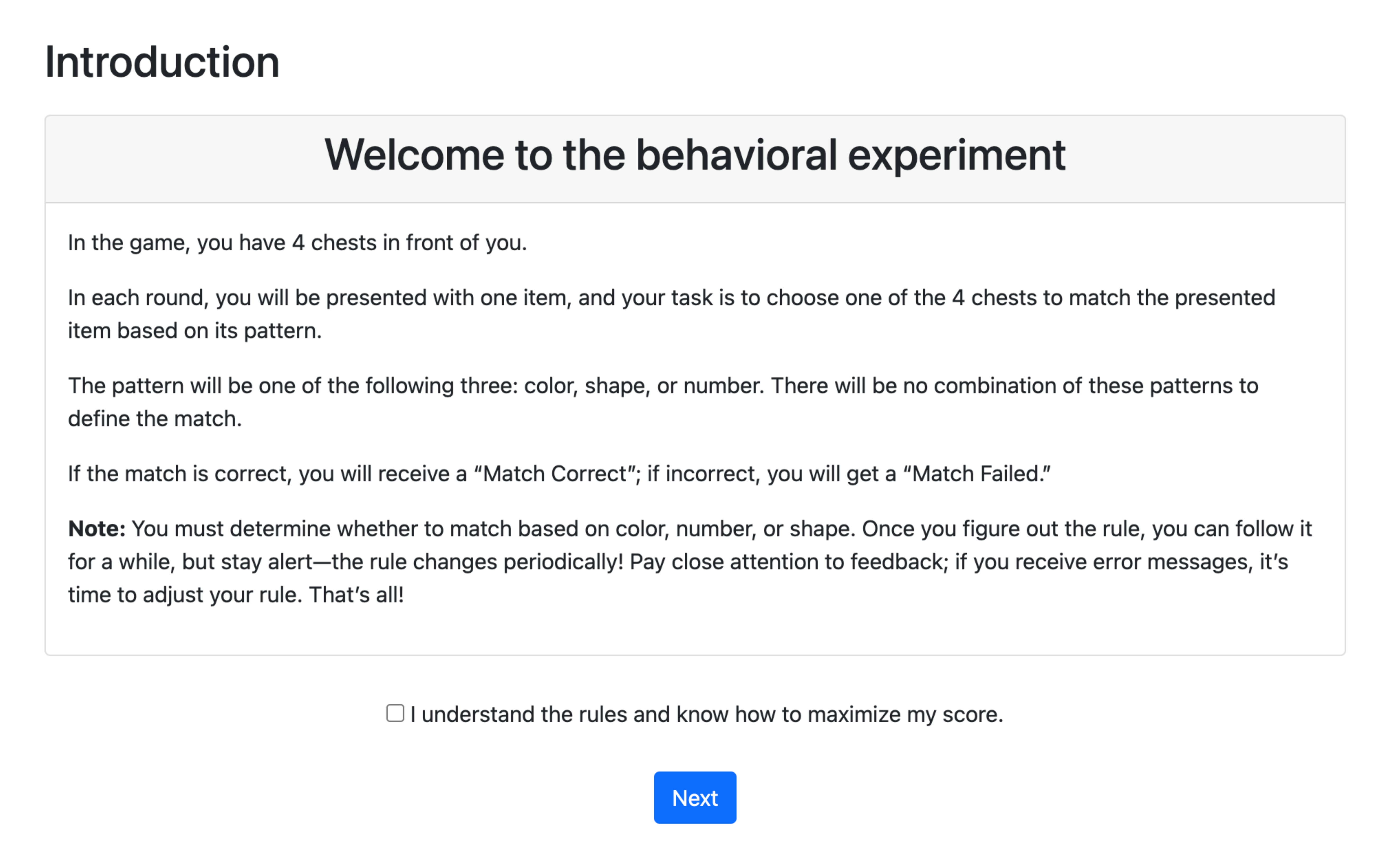}
\vspace{1em}
\caption{Pages of introduction on Wisconsin Card Sorting Task. Once participants have understood the task, they can begin by clicking the "Next" button.}
\label{fig:intro_of_wcst}
\end{figure}

\begin{figure}
\centering
\includegraphics[width=0.8\textwidth]{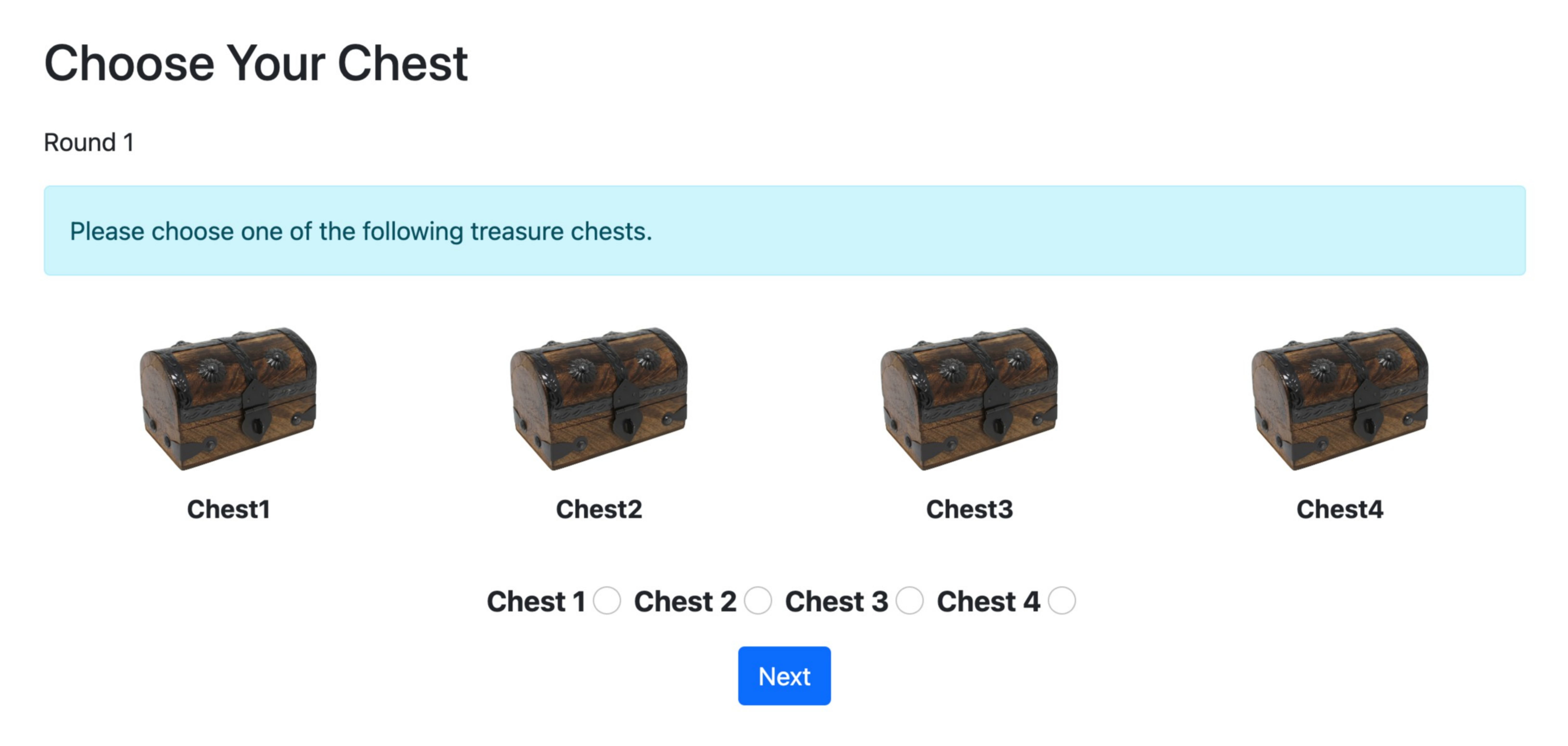}
\includegraphics[width=0.8\textwidth]{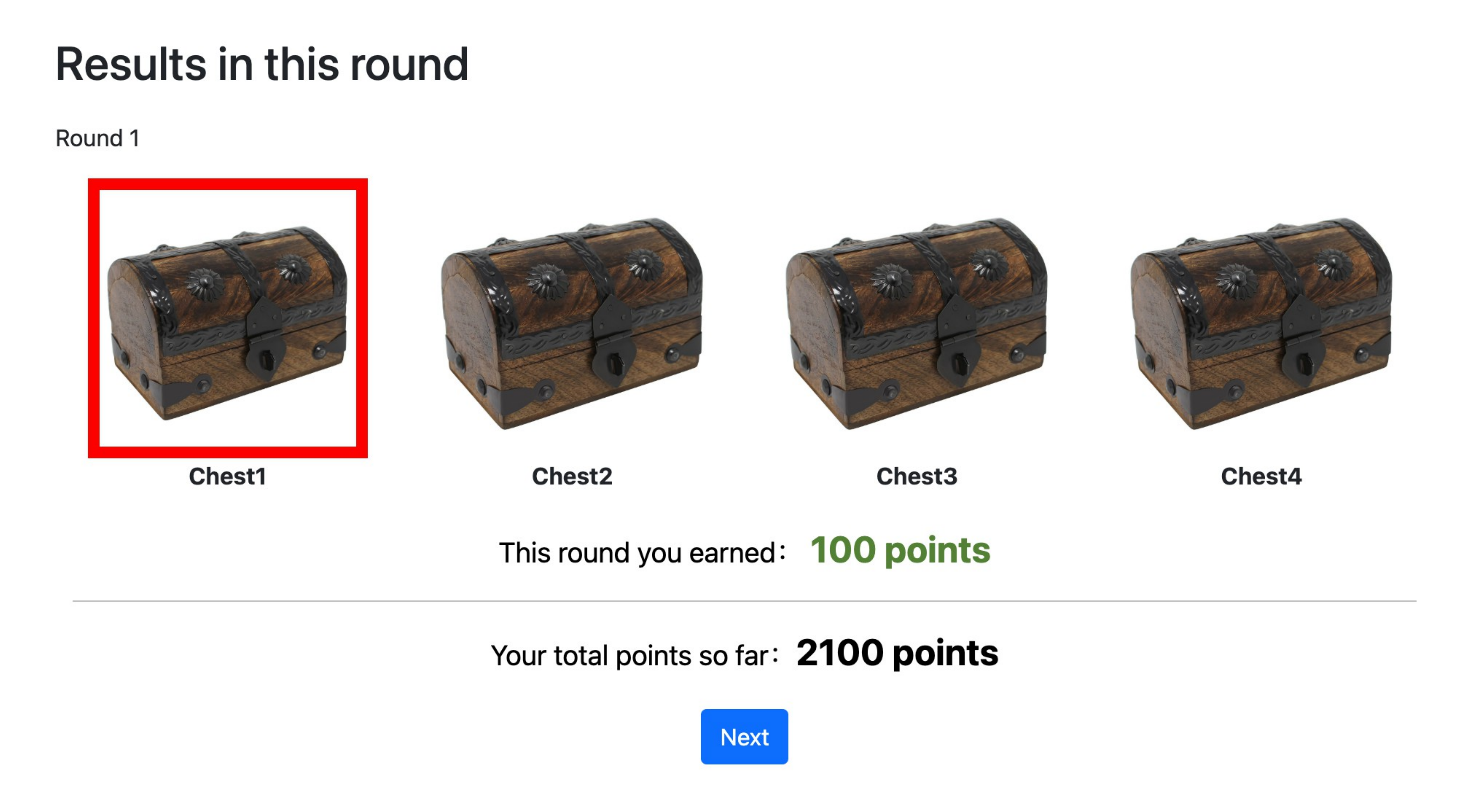}
\vspace{1em}
\caption{Choice and Result Pages on Iowa Gambling Task.}
\label{fig:choice_result_of_igt}
\end{figure}

\begin{figure}
\centering
\includegraphics[width=0.8\textwidth]{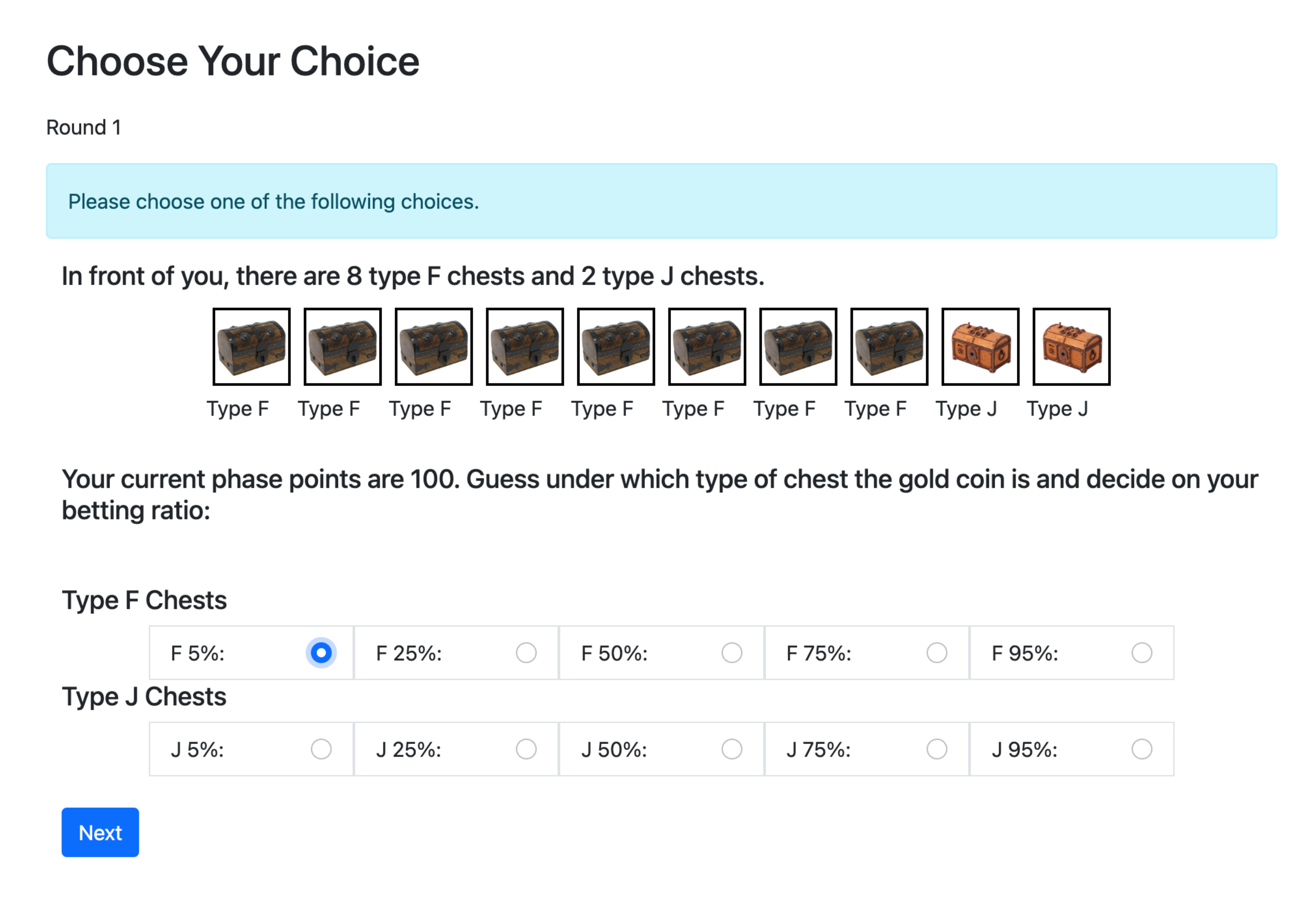}
\includegraphics[width=0.8\textwidth]{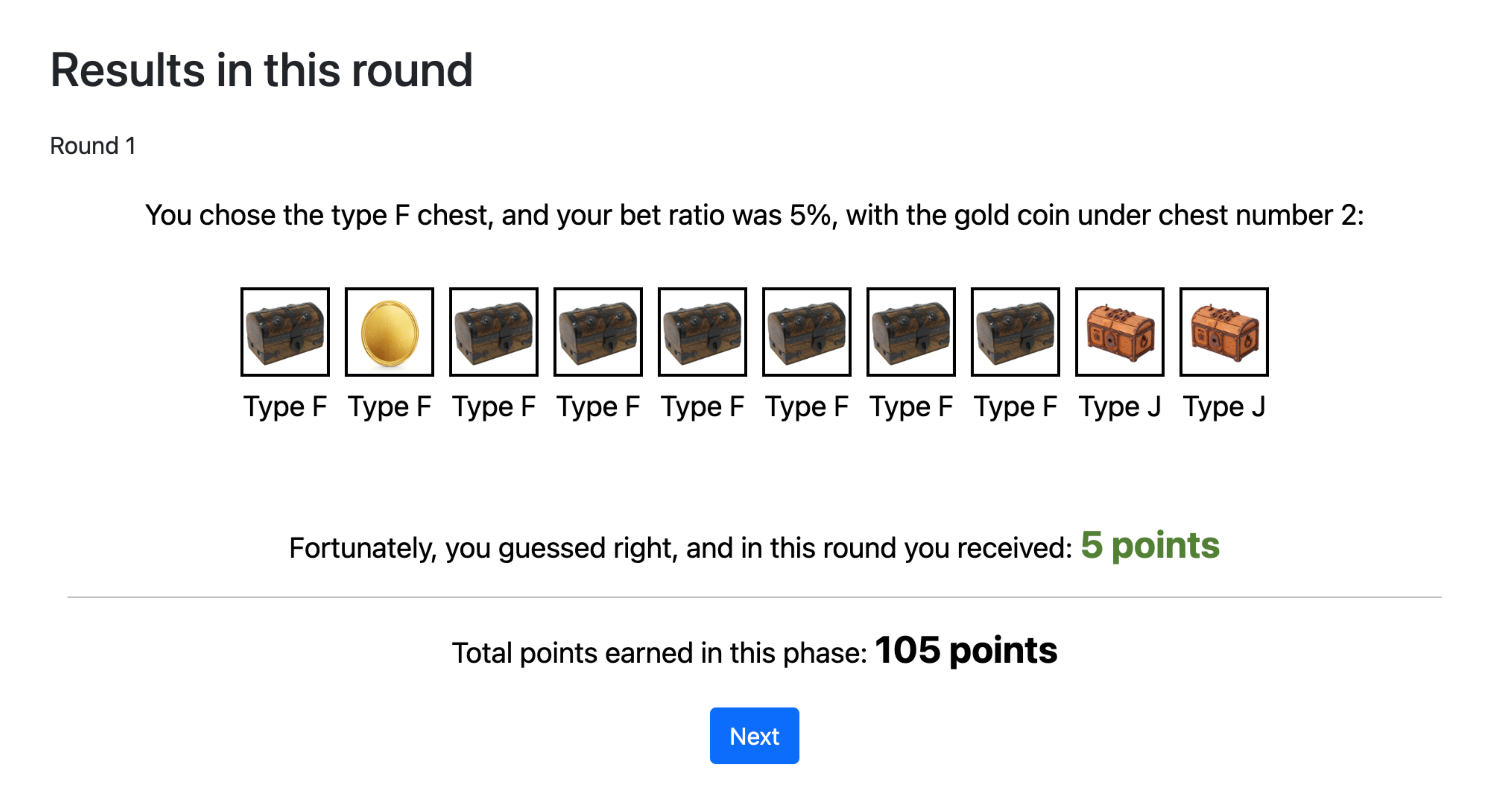}
\vspace{1em}
\caption{Choice and Result Pages on Cambridge Gambling Task.}
\label{fig:choice_result_of_cgt}
\end{figure}

\begin{figure}
\centering
\includegraphics[width=0.8\textwidth]{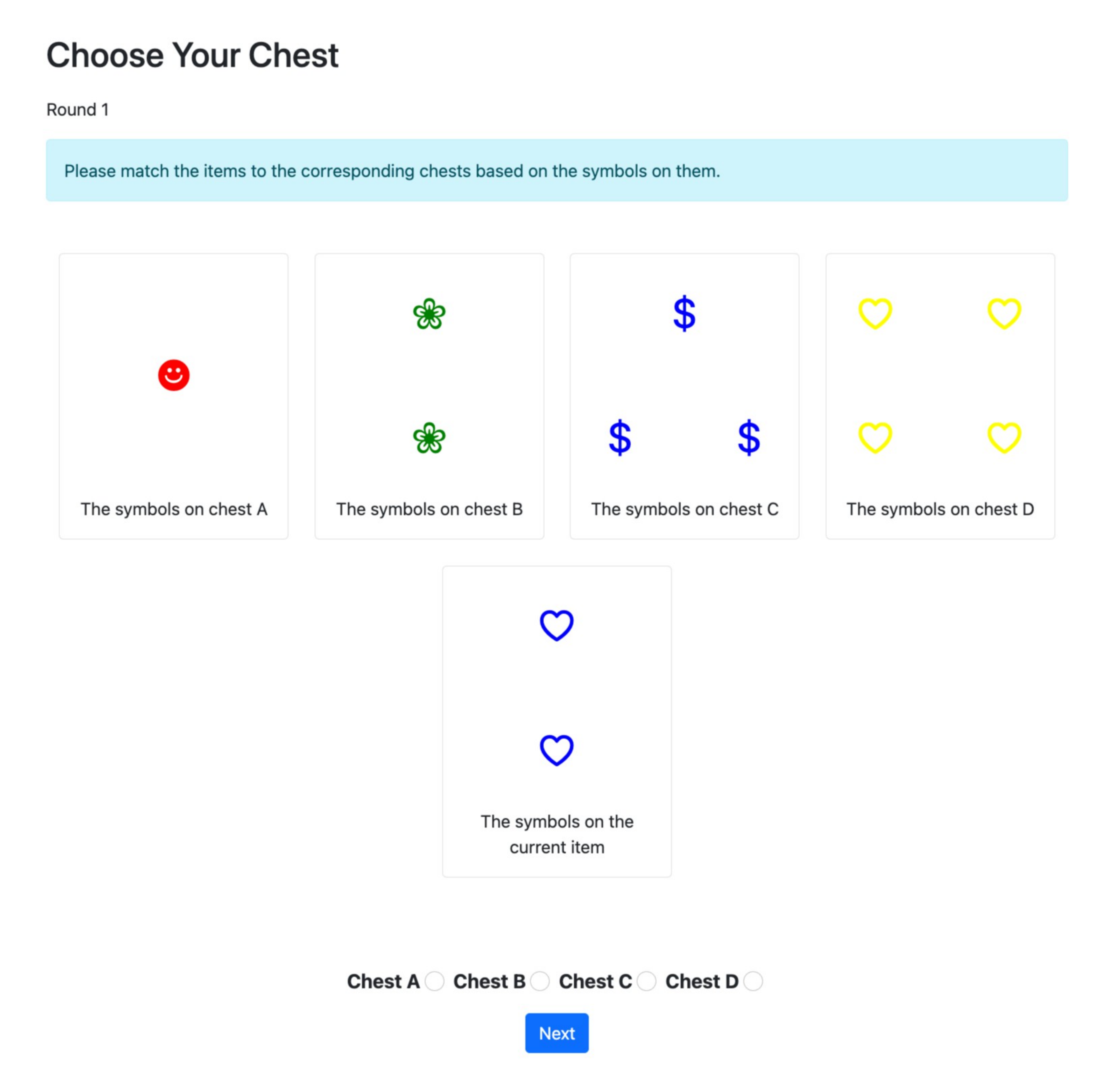}
\vspace{1em}
\caption{Choice Page on Wisconsin Card Sorting Task.}
\label{fig:choice_of_wcst}
\end{figure}

\begin{figure}
\centering
\includegraphics[width=0.8\textwidth]{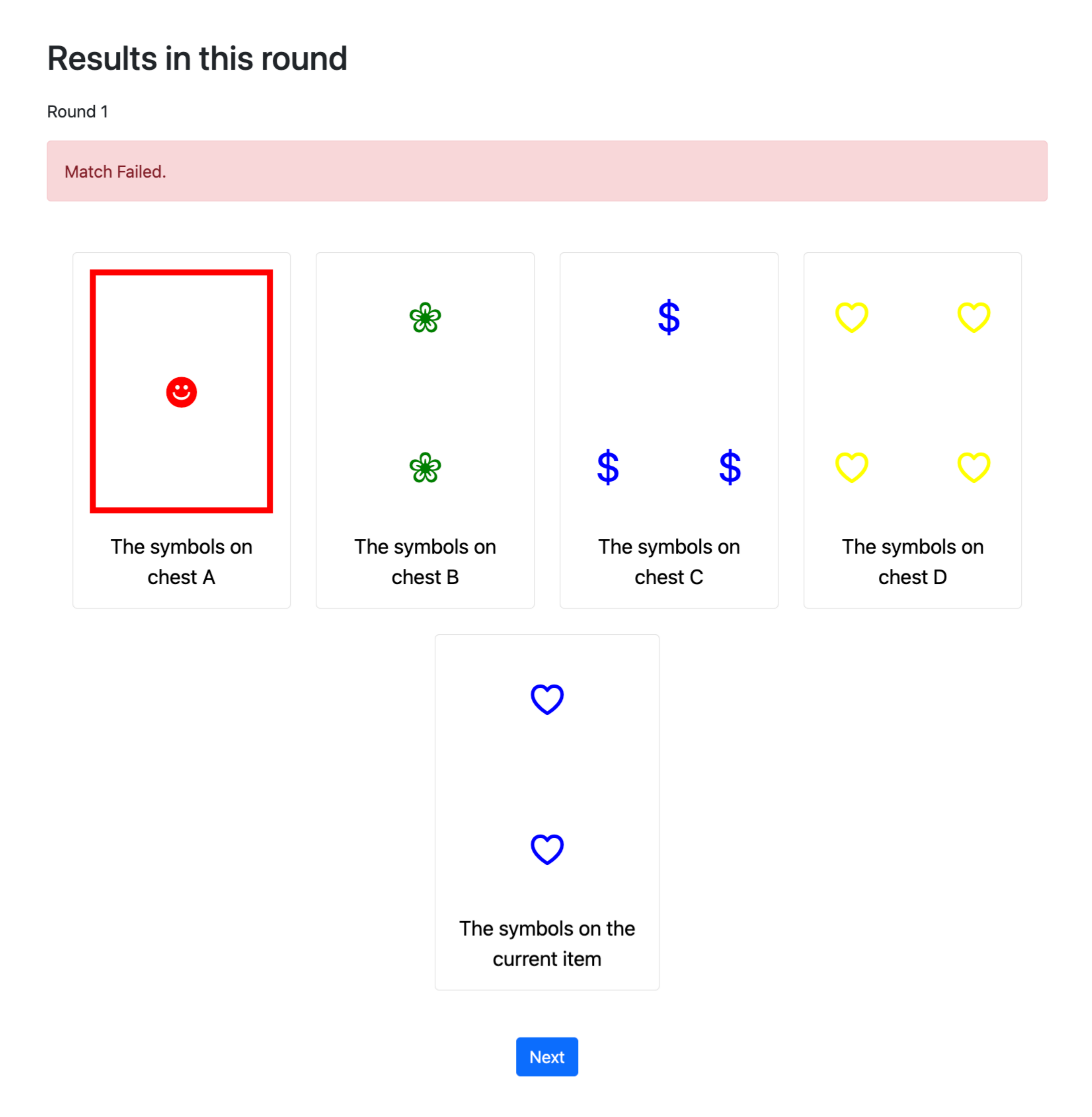}
\vspace{1em}
\caption{Result Page on Wisconsin Card Sorting Task.}
\label{fig:result_of_wcst}
\end{figure}

\begin{figure}
\centering
\includegraphics[width=0.8\textwidth]{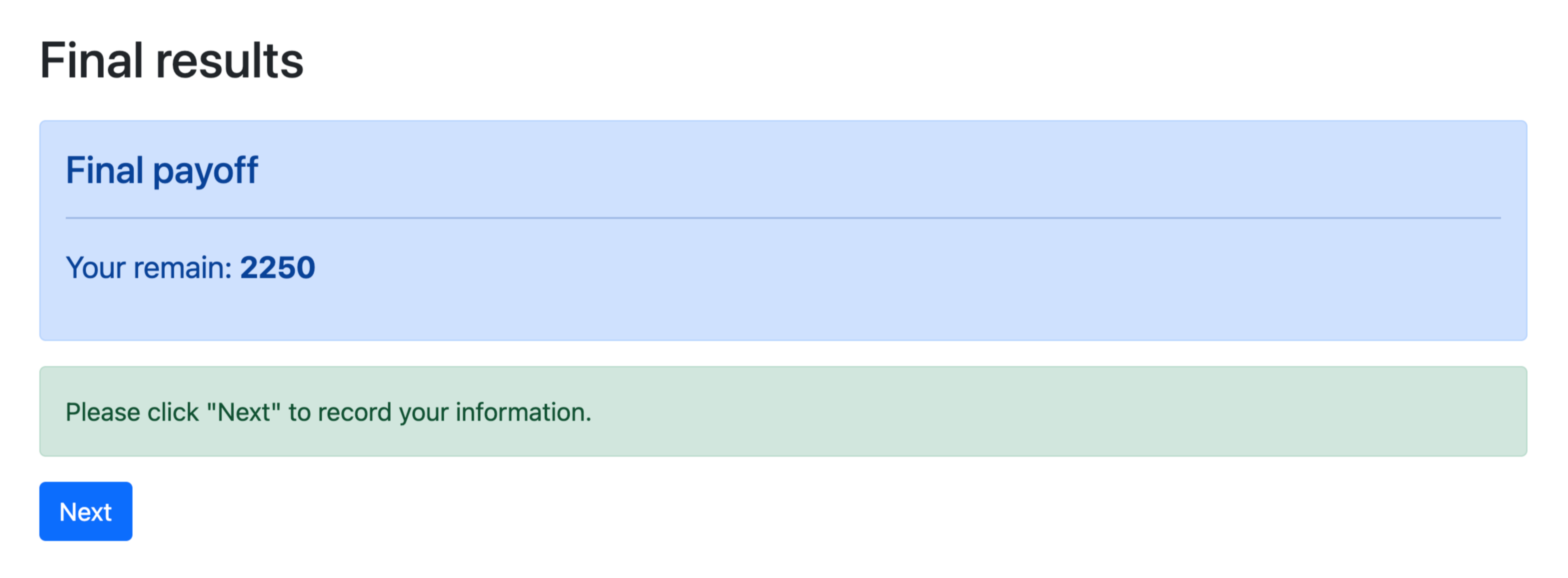}
\vspace{1em}
\caption{Final Result Page will display the participant's score for the current task.}
\label{fig:final_result}
\end{figure}

\begin{figure}
\centering
\includegraphics[width=0.75\textwidth]{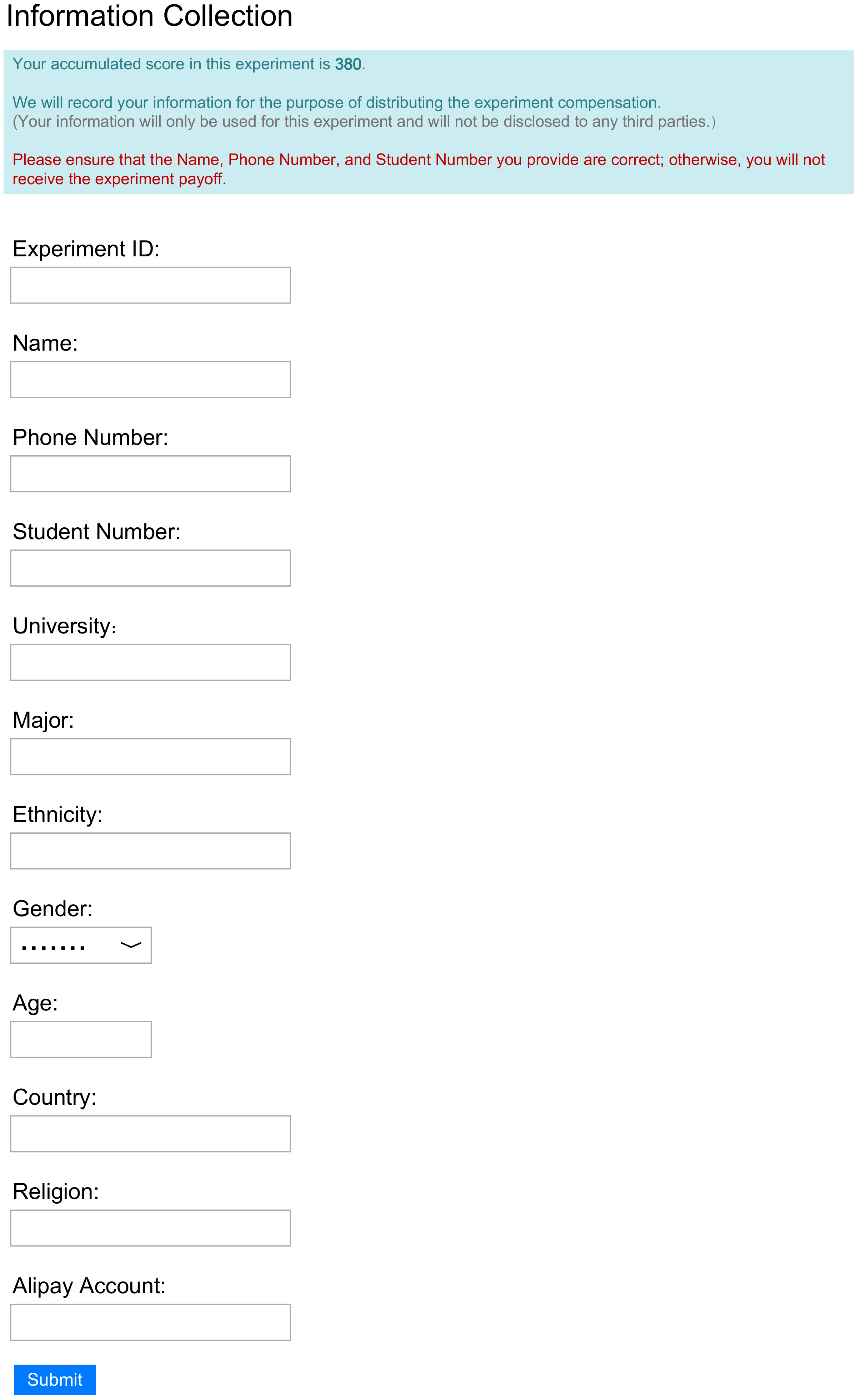}
\vspace{1em}
\caption{Demographic Information Collection Page at the end of all tasks.}
\label{fig:demographic}
\end{figure}

\begin{figure}
\centering
\includegraphics[width=0.75\textwidth]{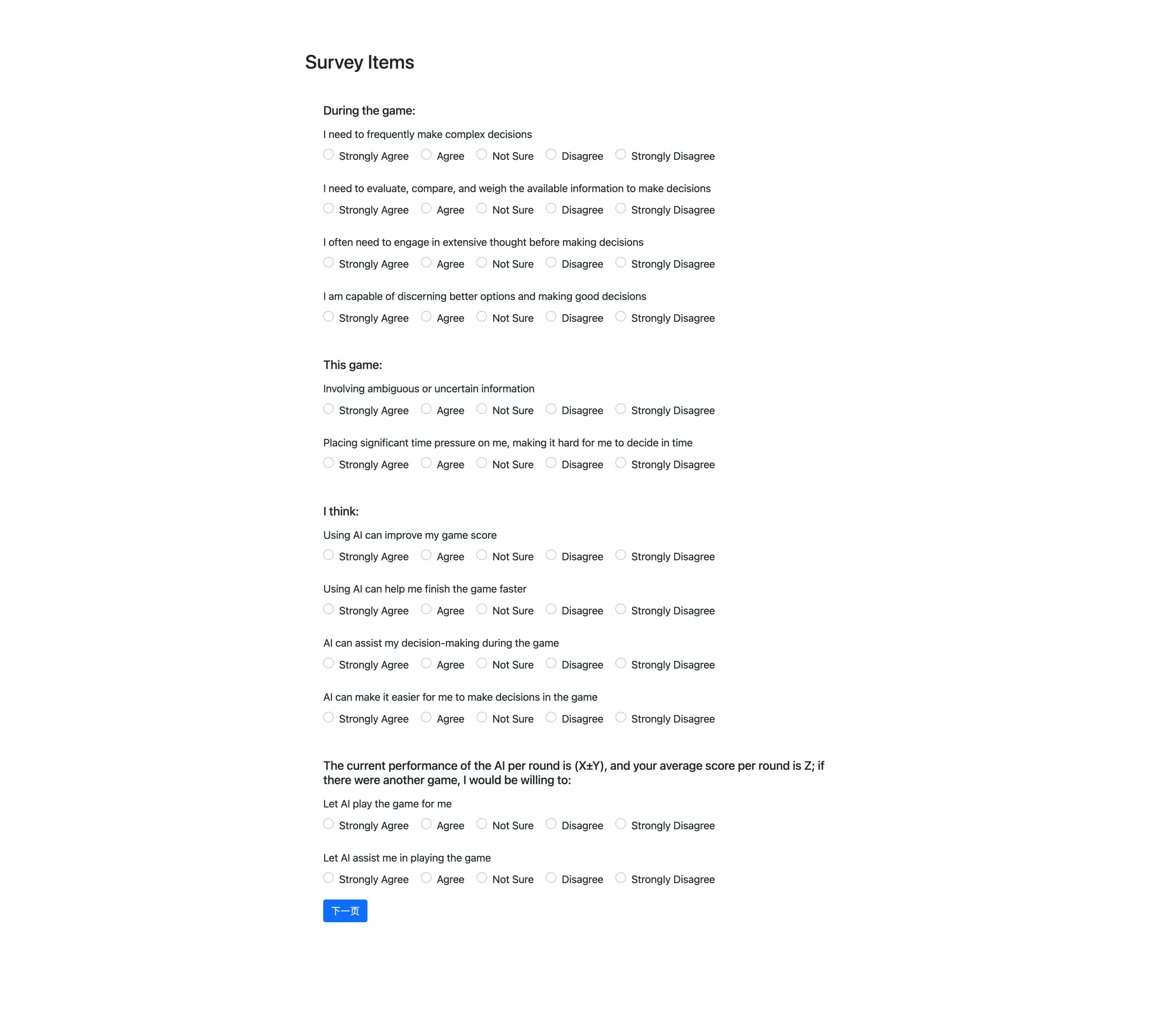}
\vspace{1em}
\caption{\textcolor{black}{Post-Task Survey on task feedback and attitudes toward AI assistance.}}
\label{fig:question}
\end{figure}

\begin{figure}
\centering
\includegraphics[width=0.8\textwidth]{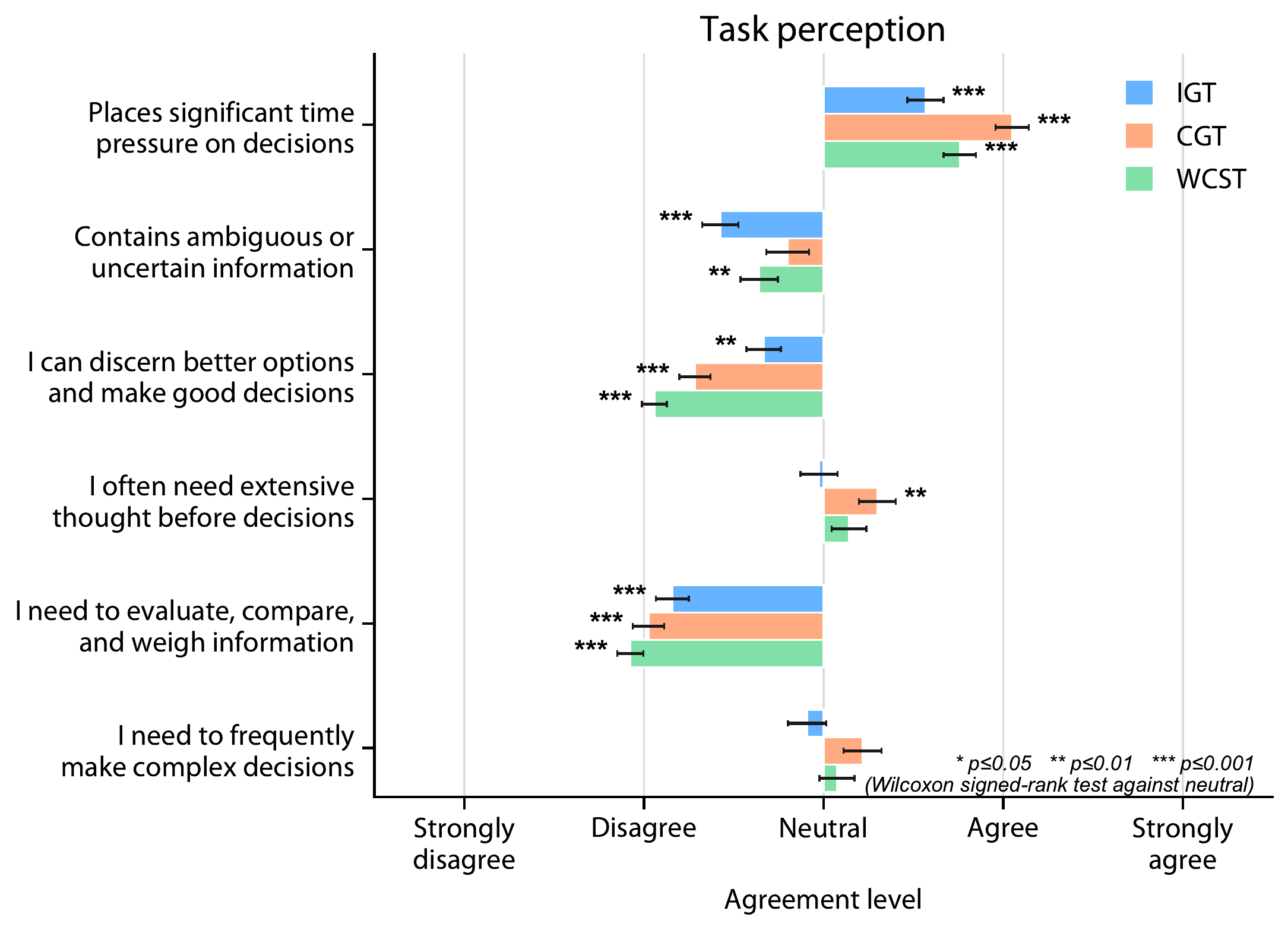}
\vspace{1em}
\caption{Participants' subjective perceptions of the three decision-making tasks. Bars represent the mean agreement level, with 95\% confidence intervals of the mean shown as error bars.}
\label{fig:task_perception}
\end{figure}


\end{appendices}



\end{document}